\documentclass[conference]{IEEEtran}

\IEEEoverridecommandlockouts 

\usepackage{amsmath}
\usepackage{mathtools}
\usepackage{amsthm}
\usepackage{amsfonts}
\usepackage{bm}
\usepackage[hidelinks]{hyperref}

\usepackage{graphicx}
\usepackage{textcomp}

\usepackage{algorithm}
\usepackage{algorithmic}
\usepackage{setspace}
\usepackage{multirow}
\usepackage{booktabs}
\usepackage[noadjust]{cite}
\usepackage{xcolor}
\usepackage{tcolorbox}
\usepackage{graphicx}

\usepackage[caption=false,font=footnotesize]{subfig}

\usepackage[capitalize,noabbrev]{cleveref}

\theoremstyle{plain}

\theoremstyle{definition}

\theoremstyle{remark}

\newcommand{\ifcomments}{\iftrue}
\definecolor{ForestGreen}{cmyk}{0.864, 0.0, 0.429, 0.396}
\definecolor{CornflowerBlue}{cmyk}{0.58,0.37,0,0.07}

\newcommand\shortsection[1]{\vspace{4pt}{\noindent\textbf{#1.}}}
\newcommand\shortersection[1]{\vspace{4pt}{\noindent\textit{#1.}}}

\usepackage{makecell}

\DeclareMathOperator*{\argmax}{argmax}
\DeclareMathOperator*{\argmin}{argmin}

\begin{document}

\title{Efficient
Semi-Supervised Adversarial Training
via Latent Clustering-Based Data Reduction}

\author{
\IEEEauthorblockN{Somrita Ghosh\textsuperscript{1} \quad Yuelin Xu\textsuperscript{2} \quad Xiao Zhang\textsuperscript{2}}
\IEEEauthorblockA{\textsuperscript{1}Independent Researcher (most research conducted while at CISPA)\\
\textsuperscript{2}CISPA Helmholtz Center for Information Security, Saarbrücken, Germany\\
\texttt{ghosh.somrita7@gmail.com}, \texttt{yuelin.xu@cispa.de},
\texttt{xiao.zhang@cispa.de}
}
\thanks{This work
has been accepted for publication at the 4th IEEE Conference on Secure and
Trustworthy Machine Learning (SaTML 2026). 
The final version will be
available on IEEE Xplore.
The implementation of our work is available as open-source code at: \url{https://github.com/TrustMLRG/EfficientSSAT}.}
}

\maketitle

\begin{abstract}
Learning robust models under adversarial settings is widely recognized as requiring a considerably large number of training samples. Recent work proposes semi-supervised adversarial training (SSAT), which utilizes external unlabeled or synthetically generated data and is currently the state of the art. However, SSAT requires substantial extra data to attain high robustness, resulting in prolonged training time and increased memory usage. In this paper, we propose data reduction strategies to improve the efficiency of SSAT by optimizing the amount of additional data incorporated. Specifically, we design novel latent clustering-based techniques to select or generate a small, critical subset of data samples near the model's decision boundary. While focusing on boundary-adjacent points, our methods maintain a balanced ratio between boundary and non-boundary data points, thereby avoiding overfitting. Comprehensive experiments across image benchmarks demonstrate that our methods can effectively reduce SSAT's data requirements and computational costs while preserving its strong robustness advantages. In particular, our latent-space selection scheme based on k-means clustering and our guided diffusion-based approach with LCG-KM are the most effective, achieving nearly identical robust accuracies with $5\times$ to $10\times$ less unlabeled data. When compared to full SSAT trained to convergence, our methods reduce total runtime by approximately $3\times$ to $4\times$ due to strategic prioritization of unlabeled data.
\end{abstract}

\section{Introduction}

Deep neural networks (DNNs) are shown to be extremely vulnerable to adversarially perturbed inputs~\cite{szegedy2013intriguing, goodfellow2014explaining}. Such a phenomenon has raised serious concerns about the reliability of deploying DNNs in security- or safety-critical applications and has driven numerous research into designing methods to enhance model robustness~\cite{goodfellow2014explaining,papernot2016distillation,buckman2018thermometer,biggio2018wild}. 
Among them, adversarial training and its variants are most popular~\cite{madry2017towards, wang2019improving, zhang2019theoretically}. 
As illustrated by Schmidt et al.~\cite{schmidt2018adversarially}, adversarially robust learning is inherently challenging, requiring a much larger sample complexity than standard learning. 
To address this, \emph{semi-supervised adversarial training} (SSAT) has been proposed to enlarge the training set size with either external unlabeled data~\cite{carmon2019unlabeled,zhai2019adversarially} or generated synthetic data~\cite{gowal2021improving,sehwag2021robust,wang2023better}.
Despite alleviating the sample complexity barrier and producing models with improved robustness performance, these SSAT algorithms typically require vast amounts of additional data, suggesting the need for much larger hardware to store the extra data and a longer training time for algorithms like adversarial training to converge.

Motivated by the above observations, we propose to study \emph{whether the large amount of extra unlabeled data is inevitable for SSAT to achieve high robust accuracy}. The ultimate goal is to maximize model robustness while utilizing as few additional data samples as possible. Inspired by Zhang et al. \cite{zhang2020geometry}, which highlights the unequal importance of training examples, we argue that with limited model capacity, SSAT should also focus on critical points near the model's decision boundary. 
Specifically, we introduce three selection strategies, including PCS, a simple approach that prioritizes data points with low prediction confidence, and two advanced latent clustering-based schemes: LCS-KM and LCS-GMM, where we identify the most uncertain region in the model's latent embedding space using clustering techniques.
The selected data points are then incorporated into the SSAT pipeline as the unlabeled dataset.
In addition, we propose a novel generative method by fine-tuning a \textit{denoising diffusion probabilistic model} (DDPM) with guidance loss inspired by our data selection schemes, including PCG, LCG-KM, and LCG-GMM, to further reduce the computation overhead of SSAT algorithms that require the pre-generation of synthetic data using diffusion models.
All our techniques are designed to be lightweight while balancing the ratio of boundary-critical points and the remaining non-boundary points, thereby reducing overall computational costs and avoiding potential overfitting to the boundary distribution.
By applying our data reduction techniques, the high requirements for unlabeled data in SSAT can be largely alleviated.

\shortsection{Contributions} 
We motivate and formalize the problem tasks of reducing the volume of unlabeled data while maintaining model robustness for SSAT algorithms (Section \ref{sec:inefficiency existing SSAT}).
To achieve this goal, we propose optimizing SSAT performance by refining the model's decision boundary in regions of high uncertainty and strategically prioritizing boundary-adjacent data points (Section \ref{sec:motivation}). 
Specifically, we introduce different selection schemes to identify boundary-adjacent points (Section \ref{sec:lcs}), including a straightforward method based on prediction confidence and two advanced selection techniques that utilize clustering of unlabeled or generated data in the model's latent embedding space. 
In addition, we propose three guided DDPM fine-tuning approaches (Section \ref{sec:lcg}) to directly generate a subset of desirable data, thereby avoiding the computational overhead of pre-generating the entire dataset.

Comprehensive experiments on image benchmarks demonstrate that our data selection strategies, particularly for the best-performing LCS-KM approach, significantly improve data efficiency compared to baseline SSAT methods, while achieving comparable test robust accuracies and reducing overall computation costs (Section \ref{sec:main results selection}). 
For instance, selecting $10\%$ extra DDPM-generated data based on LCS-KM in the SSAT algorithm can achieve robustness performance on SVHN similar to that of using the entire unlabeled dataset. Compared to full SSAT trained to convergence, this reduces the total runtime to approximately $25\%$ due to faster convergence enabled by the strategically selected data.
By directly generating the exact number of boundary-adjacent data points, our guided DDPM fine-tuning approach, such as LCG-KM, can further reduce computational overhead while maintaining robustness (Section \ref{sec:main results generation}).
Moreover, we demonstrate the generalizability of our data reduction schemes to a COVID-related medical application task (Section \ref{app:medical}) and visualize the distributions of selected data points to illustrate why our LCS-KM method consistently performs the best (Section \ref{sec:insight of our method}).
We conclude with the limitations of our study and future directions (Section \ref{sec:conclusion}).
Our work highlights the importance of prioritizing boundary-adjacent points and latent-based k-means clustering in SSAT, potentially assisting in the development of more efficient, robust learning algorithms for real-world applications.

\section{Related Work}
\label{sec:related work}

\subsection{Adversarially Robust Learning} 
\label{sec: defenses against adversarial examples}

Adversarial examples are inputs crafted with small perturbations that are designed to mislead model predictions~\cite{szegedy2013intriguing}. Earlier attacks, such as \textit{fast gradient sign method} (FGSM) \cite{goodfellow2014explaining} and \textit{projected gradient descent} (PGD) \cite{madry2017towards}, proposed gradient-based constrained optimization to find adversarial examples. Carlini and Wagner introduced CW attacks to search for minimal but unconstrained perturbations that cause misclassifications~\cite{carlini2017towards}. Recent Auto-PGD attacks \cite{pmlr-v119-croce20b} automate gradient steps to more efficiently produce worst-case perturbations, while AutoAttack \cite{croce2020robustbench} is an ensemble of diverse, parameter-free attacks, making them most reliable for evaluating the robustness of DNNs. 

\shortsection{Adversarial Training} 
To attain robustness, \textit{adversarial training} (AT) and its variants are most popular. For instance, Madry et al. proposed to train DNNs on adversarial inputs using PGD attacks~\cite{madry2017towards}, while Zhang et al. introduced TRADES~\cite{zhang2019theoretically}, which balances the trade-off between robustness and standard accuracy. 
Nevertheless, adversarial training algorithms are often criticized for overfitting to adversarially perturbed samples produced during training and for their high computational costs, which hinder their deployment in real-world applications.
In particular, Rice et al. empirically demonstrated that robust overfitting is a prevalent phenomenon in adversarial training~\cite{rice2020overfitting}, where early stopping is suggested as an effective approach for mitigating this issue. 
Techniques, such as Free AT \cite{shafahi2019adversarial} and Fast AT \cite{wong2020fast}, offer promising solutions to reduce the computational burden of adversarial training. However, they inevitably suffer from decreased robustness performance.

Moreover, a line of research has pointed out that treating all data points equally in adversarial training is not optimal, as different examples can impact model robustness differently.
For example, Zhang et al. argued that data points far from decision boundaries are inherently more secure and less likely to be influential in robust learning against adversarial perturbations~\cite{zhang2020geometry}. Consequently, they should be given less importance during training to prevent the model from over-optimization, which could negatively affect its generalization on unseen test data. In a similar vein, Hua et al. advocated for a more targeted approach~\cite{hua2021bullettrain}, recommending PGD training applied primarily to examples situated near the model's decision boundaries. The above techniques enhance the model's adversarial robustness, as perturbations on boundary-adjacent data are more informative in guiding the optimization process of adversarial training. 
Our work builds on these ideas by strategically prioritizing boundary points from the unlabeled dataset to address the unique challenges associated with SSAT algorithms.

\shortsection{SSAT} 
Utilizing additional unlabeled data to improve model performance is an active field of research~\cite{4787647}.
In the adversarial context, Schmidt et al. \cite{schmidt2018adversarially} highlighted a key observation: robust generalization often necessitates a significantly larger training set size to attain a similar level of performance compared to standard learning. 
To address the sample complexity challenge, a line of work has proposed \textit{semi-supervised adversarial training} (SSAT), which utilizes extra unlabeled data, often acquired from a similar but slightly different distribution \cite{carmon2019unlabeled, alayrac2019labels, najafi2019robustness, zhai2019adversarially}. In particular, Carmon et al. introduced the robust self-training framework~\cite{carmon2019unlabeled}, which first trains an intermediate model using available labeled data and then uses the model to generate pseudo-labels for unlabeled data. The pseudo-labeled samples, combined with the original labeled dataset, are subsequently used to train a final robust model.
Concurrently, Najafi et al. proposed an optimization framework that incorporates both labeled and unlabeled data~\cite{najafi2019robustness}, providing formal guarantees on adversarial robustness. 

Leveraging synthetic data produced by state-of-the-art generative models to build robust models has been explored in more recent adversarial ML literature \cite{gowal2021improving,sehwag2021robust,wang2023better}.
For instance, Gowal et al. employed various unconditional generative models, including DDPM~\cite{ho2020denoising}, to produce a large synthetic dataset for incorporation in adversarial training~\cite{gowal2021improving}. 
While enhancing robustness, all the aforementioned SSAT methods require a considerable amount of additional unlabeled data.
In this work, we explore data selection schemes to reduce the amount of unlabeled or generated data, aiming to achieve both efficiency and effectiveness in semi-supervised adversarial training.

\subsection{Data-Efficient Deep Learning}

Within the broader literature of data-efficient deep learning, various techniques have been proposed to reduce the amount of training data while retaining critical information. Two prominent approaches are dataset distillation and coreset selection.

\shortsection{Dataset Distillation} 
Dataset distillation aims to distill large datasets into compact, synthetic, yet informative datasets by matching gradients or optimizing specific loss functions.
For instance, Wang et al. proposed a foundational approach to dataset distillation~\cite{wang2018dataset}, condensing large datasets into smaller, synthetic subsets while preserving model performance. Later, Zhao et al. introduced the concept of gradient matching~\cite{zhao2020dataset}, where distilled data points are optimized to match the gradients of the original dataset. This technique enhances distillation performance, allowing models to learn more effectively from reduced data. In addition, Sucholutsky and Schonlau proposed soft-label dataset distillation, which uses probabilistic labels for distilled data~\cite{sucholutsky2021soft}, facilitating the training of complex models with minimal data while maintaining strong performance.
Despite its success in standard deep learning, dataset distillation is often impractical for robust learning due to its high computational demands. Moreover, it assumes that all the examples in the entire dataset are equally important for distillation, ignoring the fact that different data points may have varying influences on model performance.

\shortsection{Coreset Selection} 
Coreset selection~\cite{sener2017active,kim2022defense} aims to identify a small, representative subset of data that retains the model performance when trained on it, often relying on diversity-based metrics or optimization techniques to ensure that the selected subset is sufficiently informative.
For example, Kaushal et al. utilized diverse models to select training data subsets to reduce labeling efforts~\cite{kaushal2019learning}, while
Xia et al. introduced the concept of moderate coreset~\cite{xia2022moderate}, focusing on data points with scores near the median to construct subsets that generalize well across different scenarios based on score distributions. Similarly, Mirzasoleiman et al. proposed selecting a weighted subset of training data that approximates the full gradient by maximizing a submodular function~\cite{pmlr-v119-mirzasoleiman20a}. 

Moreover, Dolatabadi et al. adapted the coreset selection technique to adversarial contexts, offering a task-specific solution to enhance the efficiency of adversarial training~\cite{dolatabadi2023adversarial}. In particular, they proposed \textit{adversarial coreset selection}, which identifies a compact data subset by minimizing the gradient approximation error, and then utilizes it for adversarial training, with selection taking place at regular intervals.
Compared with vanilla AT, the growing reliance on large volumes of generated or external unlabeled data for SSAT presents unique challenges. Due to the significantly expanded training dataset of SSAT, iteratively refining data selection during the entire training process is computationally infeasible.
Our work complements the aforementioned methods for efficient SSAT, focusing on developing improved data reduction schemes that do not compromise robustness or scalability.

\section{Inefficiency of Existing SSAT Methods}
\label{sec:inefficiency existing SSAT}

\subsection{Semi-Supervised Adversarial Training}
\label{sec:preliminary}

We work with adversarially robust classification tasks and focus on \emph{semi-supervised adversarial training} (SSAT), which was initially introduced to the field by Carmon et al. \cite{carmon2019unlabeled}. 
Let $\mathcal{X} \subseteq \mathbb{R}^d$ be the input space and $\mathcal{Y}$ be the output space of class labels. Assume \( D_l \) is the underlying labeled data distribution over $\mathcal{X} \times \mathcal{Y}$ that we aim to learn adversarially robust classifiers. 
Let \( \mathcal{S}_l \) be the training dataset with examples independently and identically drawn from \( D_l \), and \( \mathcal{S}_u \) be a collection of inputs sampled from an unlabeled data distribution \( D_u \) defined over $\mathcal{X}$. 
In adversarial ML literature, \( D_u \) is typically considered to be similar but not identical to the marginal input distribution of \( D_l \), and the set size of $\mathcal{S}_u$, denoted as \( |\mathcal{S}_u| \), to be much larger than \( |\mathcal{S}_l| \), since unlabeled data are much easier to acquire than well-annotated labeled data. For instance, when $\mathcal{S}_l$ corresponds to the $50$K labeled CIFAR-10 training images, 
$\mathcal{S}_u$ is the $500$K ``most CIFAR-10-like but non-identical'' images selected from the whole $80$M Tiny ImageNet dataset~\cite{carmon2019unlabeled}. 

Specifically, SSAT usually first trains an intermediate model $f_{\hat\theta}$ on the labeled training set \( \mathcal{S}_l \), known as the \textit{pseudo labeler}, and then assigns pseudo-labels to each unlabeled data point \( \bm{x}\in\mathcal{S}_u \). The weight parameters of the intermediate model are optimized by minimizing the following objective function:
\begin{align}
\label{eq:pseudo labeler}
    \hat\theta = \argmin_{\theta} \ \frac{1}{|\mathcal{S}_l|} \sum_{(\bm{x}, y)\in\mathcal{S}_l} \ell(\theta, \bm{x}, y),
\end{align}
where $\ell(\cdot)$ stands for the individual standard loss (e.g., cross-entropy), capturing the discrepancy between model predictions and the ground-truth labels.
Finally, SSAT adversarially trains a classification model, denoted as $\mathrm{SSAT}(\mathcal{S}_l, \mathcal{S}_u, \gamma)$, on both labeled data $\mathcal{S}_l$ and unlabeled data \( \mathcal{S}_u \) but paired with pseudo labels by minimizing the following adversarial loss:
\begin{align}
\label{eq:SSAT}
\mathrm{SSAT}(\mathcal{S}_l, \mathcal{S}_u, &\gamma) = \argmin_{\theta} \quad \frac{\gamma}{|\mathcal{S}_l|} \sum_{(\bm{x}, y) \in \mathcal{S}_l} 
   \ell_{\mathrm{adv}} (\theta, \bm{x}, y) \nonumber \\
 & + \frac{1 - \gamma}{|\mathcal{S}_u|}  
   \sum_{\bm{x} \in \mathcal{S}_u} 
   \ell_{\mathrm{adv}} \!\left( \theta, \bm{x}, f_{\hat{\theta}}(\bm{x}) \right),
\end{align}
where $\gamma \geq 0$ is a hyperparameter that controls the trade-off between labeled and unlabeled data. 
A typical approach used in prior work is to construct each training batch with varying ratios of labeled and unlabeled data (corresponding to different values of $\gamma$) but fix the batch size to optimize for the best SSAT performance~\cite{gowal2021improving}.
For simplicity, we term $\gamma$ as the \emph{per-batch ratio}.
When $\gamma$ is set as $1$, Equation \ref{eq:SSAT} reduces to the optimization objective of vanilla \emph{adversarial training} (AT) \cite{madry2017towards}.
In Equation \ref{eq:SSAT}, $\ell_{\mathrm{adv}}(\cdot)$ denotes the individual adversarial loss. Specifically, for any $(\bm{x}, y)\in\mathcal{S}_l$ and strength $\epsilon>0$,
\begin{align}
\label{eq:adv loss}
    \ell_{\mathrm{adv}} (\theta, \bm{x}, y) = \max_{\bm{\delta} \in \mathcal{B}_\epsilon(\bm{0})} \ell(\theta, \bm{x} + \bm{\delta}, y), 
\end{align}
where $\mathcal{B}_\epsilon(\bm{0})$ denotes the perturbation ball centered at $\bm{0}$ with radius $\epsilon$ in some distance metric such as $\ell_p$-norm.
Aligned with prior literature~\cite{madry2017towards}, we adopt multi-step PGD to approximately solve for the optimization problem in Equation \ref{eq:adv loss}.

\subsection{Limitation of Prior Work} 
\label{sec: limitation}

\shortsection{Data Inefficiency}
Although incorporating unlabeled data alleviates the sample complexity barriers of adversarially robust learning~\cite{schmidt2018adversarially} and can produce models with higher robust accuracies, they require
a substantial amount of extra unlabeled data $\mathcal{S}_u$ to ensure effective robustness enhancement.
Such a trend is clearly documented in the leaderboard of RobustBench \cite{croce2020robustbench}.
For CIFAR-10 image classification tasks, state-of-the-art SSAT algorithms either select $500$K external Tiny ImageNet samples or generate millions of synthetic CIFAR-10-like images, both significantly larger than the original $50$K CIFAR-10 training images.
Specifically, the method proposed by Gowal et al.~\cite{gowal2021improving} achieves around $65\%$ robust accuracy on CIFAR-10 using WideResNet-70-16 against $\ell_\infty$ perturbations with radius $\epsilon=8/255$, increasing the robustness performance of vanilla AT~\cite{madry2017towards} by a large margin; however, it relies on an extra unlabeled set of $100$M DDPM-generated data.
Training SSAT algorithms on the vast, unlabeled dataset requires significantly more computing power, such as specialized hardware, data parallelism, or distributed training, which in turn induces much higher energy consumption and a larger carbon footprint.
In addition, handling a larger dataset in machine learning often means increased GPU processing load and memory usage.

\shortsection{High Computation Cost}
Apart from the data inefficiency, SSAT requires a longer convergence time to obtain the best-performing model, usually $2$ to $4$ times training epochs compared with vanilla AT (see Figure \ref{fig:convergence} for supporting evidence). 
Intuitively, learning meaningful, robust representations through adversarial training from such a large and potentially diverse unlabeled dataset is more challenging.
As will be illustrated in our experiments, the slower convergence of SSAT can be attributed mainly to the large set of unlabeled data (paired with pseudo labels), which often induces higher gradient variance than the original labeled data samples.
Once we reduce the size of the unlabeled data while retaining the essential information, SSAT algorithms are observed to converge much faster.
We note that vanilla AT has already been criticized for its high computation costs~\cite{shafahi2019adversarial,wong2020fast}, due to the iterative PGD steps involved in solving the inner maximization problem (Equation \ref{eq:adv loss}). SSAT algorithms incur even higher computations, as they converge more slowly and often employ larger model architectures (see Table \ref{table:SVHN+CIFAR10} for concrete runtime comparisons). 
Therefore, developing more efficient methods, especially for resource-constrained scenarios, becomes critical for SSAT to effectively learn the core information in the unlabeled dataset.

\subsection{Problem Formulation}
\label{sec:problem formulation}

Recognizing the data and computational inefficiencies of SSAT, we propose investigating whether the large size of unlabeled data is necessary for achieving high model robustness.
To be more specific, we introduce two general methodology frameworks for improving the efficiency of SSAT algorithms, corresponding to the following optimization problems:

\shortsection{Strategic Selection}
Inspired by the idea of coreset selection for data-efficient deep learning \cite{kaushal2019learning, xia2022moderate, pmlr-v119-mirzasoleiman20a}, the first approach aims to strategically search for a small but essential set of unlabeled data $\mathcal{A}_u \subseteq \mathcal{S}_u$ such that SSAT with the selected subset $\mathcal{A}_u$ can produce models with comparable robustness to those obtained using the full unlabeled dataset $\mathcal{S}_u$.
Formally, we aim to solve the following constrained optimization problem:
\begin{align}
\label{eq:problem formulation}
    \max_{\mathcal{A}_u \subseteq \mathcal{S}_u} \mathrm{AdvRob}_\epsilon \big(\mathrm{SSAT}(\mathcal{S}_l, \mathcal{A}_u, \gamma)\big), \text{ s.t. } |\mathcal{A}_u| \leq \alpha |\mathcal{S}_u|,
\end{align}
where $\mathrm{SSAT}(\mathcal{S}_l, \mathcal{A}_u, \gamma)$ denotes the model learned by SSAT with $\mathcal{S}_l$ and $\mathcal{A}_u$ based on Equation \ref{eq:SSAT}, and $\alpha\in(0,1)$ is a predefined ratio capturing the data constraint. Here, $\mathrm{AdvRob}_\epsilon(\theta)$ denotes the model robustness on $\mathcal{D}_l$ under $\epsilon$ perturbations:
\begin{align}
\label{eq:def adv rob}
    \mathrm{AdvRob}_\epsilon(\theta)\!=\!1\!-\!\mathbb{E}_{(\bm{x},y)\sim D_l}\!\left[\max_{\bm{\delta}\in\mathcal{B}_\epsilon(\bm{0})}\!\ell_{0/1}(\theta,\bm{x}+\bm{\delta},y)\right],
\end{align}
where $\ell_{0/1}$ denotes the 0-1 loss function. 
Due to the combinatorial nature and the high computation costs of AT algorithms, it is computationally hard to enumerate all the feasible subsets $\mathcal{A}_u$ to solve the proposed optimization problem exactly.
As we will illustrate in Section \ref{sec:lcs}, we design different selection schemes for the extra unlabeled dataset $\mathcal{S}_u$ that are effective in approximately solving the optimization problem in Equation \ref{eq:problem formulation} while only incurring negligible computational overhead.

\shortsection{Guided Diffusion}
The second method proposes to leverage existing pretrained diffusion models but fine-tune them to directly generate a small set of extra data with desirable properties, ready to be incorporated into the SSAT pipeline.
Similar to our first strategic selection approach, we aim to achieve comparable robustness to existing SSAT while maximizing the efficiency for the utilization of the additional data:
\begin{align}
\label{eq:problem formulation generation}
    \max_{\mathcal{G}_u} \mathrm{AdvRob}_\epsilon \big(\mathrm{SSAT}(\mathcal{S}_l, \mathcal{G}_u, \gamma)\big), \text{ s.t. } |\mathcal{G}_u| \leq \alpha |\mathcal{S}_u|,
\end{align}
where $\mathcal{G}_u$ denotes a small set of unlabeled data generated by a fine-tuned generative model. As will be detailed in Section \ref{sec:lcg}, we adopt a pretrained DDPM~\cite{ho2020denoising} as the backbone while strategically fine-tuning it with novel training losses for guided data generations.
Compared with the previous selection-based method, a guided diffusion approach can avoid the inefficient pre-generation of a large amount of unlabeled data, making it particularly suitable for SSAT methods that already utilize a pre-trained diffusion model for synthetic data generation.

\section{Improving the Efficiency of SSAT via Latent Clustering-Based Data Reduction}
\label{sec:method}

So far, we have formulated the problem tasks of reducing the unlabeled dataset size to improve SSAT's efficiency while maximizing the robustness enhancement. 
In this section, we first explain why boundary-adjacent data points are essential for more efficient utilization of unlabeled data (Section \ref{sec:motivation}), then we detail the proposed method designs for both strategic selection and guided generation (Sections \ref{sec:lcs} and \ref{sec:lcg}).

\subsection{Prioritize Boundary-Adjacent Unlabeled Data}
\label{sec:motivation}

Achieving high robustness with restricted data resources remains challenging, as it involves striking a balance between data significance, optimizing the use of unlabeled data, and tackling constraints imposed by model capacity.
Inspired by prior literature~\cite{zhang2020geometry, hua2021bullettrain} that emphasizes the unbalanced data importance for adversarial training, we hypothesize that not all unlabeled data contribute equally to the robustness enhancement for SSAT. 
In particular, we propose identifying or generating a small set of \textit{vulnerable yet valuable} unlabeled data points that are close to the model's decision boundary. 
Consequently, such boundary-adjacent data points are highly susceptible to label changes under small input perturbations and are inherently difficult for the model to classify.
Thus, improving their classification can yield enhanced robustness against adversarial inputs.
We expect the decision boundary of the intermediate model $f_{\hat\theta}$ to provide a meaningful proxy for locating difficult-to-classify data points.
Since $f_{\hat\theta}$ is usually trained with strong standard accuracy, it is expected to preserve class semantics and provide useful information for identifying these critical points.
By focusing more on these critical boundary points, the final model can be trained more efficiently using SSAT algorithms while upholding robust accuracies.

While a plausible approach is to select a set of unlabeled data based on how much their label predictions change under adversarial perturbations returned by PGD attacks, it is computationally expensive, undermining the efficiency gains we aim for. 
In particular, such a PGD-based method entails iterative optimization processes to pinpoint boundary points, making them impractical for large-scale unlabeled datasets. 
Therefore, we seek better alternatives to effectively identify boundary data points without incurring high computational overhead.

\subsection{Strategic Selection via Latent Clustering}
\label{sec:lcs}

\shortsection{PCS}
To identify a core subset of boundary-adjacent points while accounting for computational efficiency, we first propose a straightforward approach, coined \emph{prediction confidence-based selection} (PCS), which makes use of the intermediate model $f_{\hat\theta}$ to compute a prediction confidence score for each unlabeled data point.
Algorithm \ref{alg:dataselection_confidence} in Appendix \ref{append:alg} presents the pseudocode of this selection scheme.
Initially, all the unlabeled data points in $\mathcal{S}_u$ are sorted by their prediction confidence $\mathrm{Conf}(\cdot)$, with those exhibiting the lowest confidence scores being prioritized. The underlying assumption is that data points with low confidence scores are more likely to lie near the model's decision boundary, making them ideal candidates for our selection. Note that the parameter $\beta\in[0,1]$ is introduced to balance the ratio between boundary and non-boundary points to avoid overfitting, which is used in all our proposed selection schemes. The reason for involving such a trade-off parameter will be further discussed in Section \ref{sec:hyperparameter tuning}.

The biggest advantage of PCS is its high computational efficiency for ranking the unlabeled data.
Note that for SSAT, we usually have an order-wise larger collection of unlabeled data than the original labeled dataset.
Thus, an efficient selection scheme is desirable to avoid high computational overhead.
Nevertheless, our experiments showed that using prediction confidence may not capture the underlying structure of the data well, leading to decreased robustness enhancement when incorporated in SSAT.
We hypothesize that PCS overlooks the geometric relationships and distributional properties, which can be crucial for characterizing boundary-adjacent points. In addition, DNNs have been shown to be overconfident in their predictions \cite{guo2017calibration}, suggesting that prediction confidence scores returned by neural network models might be inherently biased.

\shortsection{LCS}
To overcome the aforementioned issues, we introduce \emph{latent clustering-based selection} (LCS) schemes to pinpoint desirable data in the model's latent space using different clustering techniques. Our approach begins by generating latent embeddings for all $\bm{z} = h_{\hat\theta}(\bm{x})$ with $\bm{x}\in\mathcal{S}_u$, where \( h_{\hat\theta} \) denotes the mapping of the input layer to the penultimate layer with respect to the intermediate model \( f_{\hat\theta} \).
The penultimate layer captures more abstract, high-level features and better represents the underlying data structure, which helps identify points near decision boundaries. It avoids the biases of overconfident predictions from the final layer, offering more reliable clustering.
Here, the goal is to identify boundary-adjacent data points inferred by examining distances to cluster centroids. Points equidistant from multiple centroids are expected to be closer to decision boundaries in the latent embedding space. 

In particular, we explore two classical clustering techniques for their unique benefits in identifying boundary points in the LCS framework. 
Algorithm \ref{alg:dataselection_clustering} depicts the pseudocode of the methods. In particular, \emph{LCS with k-means} (LCS-KM) generates latent representations of unlabeled data and clusters them based on Euclidean distances to the centroids, prioritizing data points that are equidistant from multiple centroids to effectively capture local geometric structures around decision boundaries. On the other hand, \emph{LCS with Gaussian mixture models} (LCS-GMM) fits the latent representations to Gaussian mixture models, using posterior probabilities across multiple fitted Gaussians to identify points that are likely near decision boundaries. Compared to PCS, both LCS-KM and LCS-GMM leverage latent space clustering to provide a more accurate characterization of boundary vulnerabilities.

\shortsection{LCS-KM} In this variant of LCS, we first generate latent embeddings \( \{ \bm{z} = h_{\hat\theta}(\bm{x}): \bm{x} \in \mathcal{S}_u \} \), and then partition the unlabeled dataset into \( k \) clusters \( \{ \mathcal{C}_1, \mathcal{C}_2, \ldots, \mathcal{C}_k \} \) by minimizing the within-cluster sum of squares \( \sum_{j=1}^k \sum_{\bm{z} \in \mathcal{C}_j} \| \bm{z} - \bm{\mu}_j \|^2 \), where \( \bm{\mu}_j \) is the centroid of the \( j \)-th cluster. For each \( \bm{z} \), we compute the Euclidean distance to each cluster centroid \( \| \bm{z} - \bm{\mu}_j \| \). Data points are selected based on the minimal difference in distance between each latent embedding $\bm{z}$ to the corresponding two closest cluster centroids \( \Delta d = |d_{1} - d_{2}| \), where \( d_{1} \) and \( d_{2} \) are the Euclidean distances to the closest and second closest centroids, respectively. Finally, the top \( \alpha |\mathcal{S}_u| \) points from \( \mathcal{S}_u \) with the smallest \( \Delta d \) values form the reduced unlabeled dataset, followed by the same balancing step using the ratio parameter $\beta$. As will be demonstrated in our experiments, prioritizing these strategically selected unlabeled data points during semi-supervised adversarial training can achieve comparable robustness with much-improved efficiency.

\shortsection{LCS-GMM} In this variant of LCS, we again start by computing latent embeddings for unlabeled data \( \{ \bm{z} = h_{\hat\theta}(\bm{x}): \bm{x} \in \mathcal{S}_u \} \). Instead of using k-means clustering, we fit these latent representations using Gaussian mixture models (GMMs). A GMM assumes that the data is generated from a mixture of \( k \) Gaussian distributions, each with its own mean \( \bm{\mu}_j \) and covariance matrix \( \mathbf{\Sigma}_j \) for the \( j \)-th component. 
Mathematically, each data point \( \bm{z} \) has a probability of belonging to the \( j \)-th Gaussian component given by the posterior probability:
\begin{align}
\label{eq:GMM posterior probability}
    p_{j}(\bm{z}) = \frac{\pi_j \cdot \mathcal{N}(\bm{z} \mid \bm{\mu}_j, \mathbf{\Sigma}_j)}{\sum_{i=1}^{k} \pi_i \cdot \mathcal{N}(\bm{z} \mid \bm{\mu}_i, \mathbf{\Sigma}_i)},
\end{align}
where \( \pi_j \) is the mixing coefficient for the \( j \)-th Gaussian component, and \( \mathcal{N}(\bm{z} \mid \bm{\mu}_j, \mathbf{\Sigma}_j) \) denotes the Gaussian distribution with mean \( \bm{\mu}_j \) and covariance \( \Sigma_j \). 
We focus on data points with similar top posterior probabilities, suggesting that they are near the boundary between Gaussian distributions. Specifically, for each \( \bm{z} \), we calculate the difference between the highest and the second-highest posterior probabilities: $\Delta p = |p_1(\bm{z}) - p_2(\bm{z})|$,
where \( p_1(\bm{z}) \) and \( p_2(\bm{z}) \) are the highest and second-highest posterior probabilities, respectively. Data points with the smallest \( \Delta \gamma \) values are selected, as they are more likely to be near decision boundaries. Finally, the top \( \alpha |\mathcal{S}_u| \) unlabeled data with the smallest \( \Delta p \) values are used to form the selected subset.
As will be illustrated in Section \ref{sec:experiment}, the robustness performance of SSAT can be largely maintained when using  LCS-KM and LCS-GMM to select a small subset of unlabeled data (e.g., $\alpha = 20\%$).
More detailed discussions of the difference between the proposed data selection schemes are provided in Section \ref{sec:insight of our method}, where we visualize the selected unlabeled data in a two-dimensional latent space.

\subsection{Guided Diffusion via Latent Clustering}
\label{sec:lcg}

While our selection strategy can effectively reduce the unlabeled data set size and lower the training time, it has a critical drawback: when SSAT employs a generative approach, all our previous methods require generating the full synthetic unlabeled dataset as a prerequisite before selection, which is inefficient in terms of both generation time and storage requirement.
As shown in Table \ref{tab:pipeline_time}, generating 1M CIFAR-10-like synthetic images via DDPM requires around $4$ hours on a single A100 GPU.
Note that the best current SSAT method, as documented in the RobustBench leaderboard~\cite{croce2020robustbench}, utilizes a few hundred million synthetically generated images, suggesting that the total generation time can be a significant computational overhead.
Therefore, a natural question arises: \emph{whether pre-generating synthetic data and then applying strategic selection schemes can be further improved.} 

In this section, we introduce an alternative data generative pipeline, detailed in Algorithm \ref{alg:diffusion_finetune}, which answers the above question affirmatively.
In particular, we fine-tune a pre-trained DDPM using novel guidance losses to achieve more efficient unlabeled data generation, building on the similar insights presented in the previous sections. In this way, we directly generate a small, critical set of boundary-adjacent data points, thereby avoiding redundant data generation and further reducing the overall computation overhead.


\shortsection{DDPM} 
Diffusion models represent a milestone in the history of generative modeling, achieving state-of-the-art performance for high-quality image generation and synthesis.
A \emph{denoising diffusion probabilistic model} (DDPM) is a deep neural network trained to learn how to step-by-step reverse the forward process of adding Gaussian noise to an image~\cite{ho2020denoising}.
By gradually removing Gaussian noise through the reverse process, DDPM can thus generate real-looking images from pure noise.
Let $q_0$ be a real image distribution over $\mathbb{R}^d$ and $\mathcal{N}(\bm{0}, \mathbf{I})$ be the noise distribution. The forward process defines how to map images $\bm{x}_0\sim q_0$ to Gaussian noises $\bm{x}_T\sim\mathcal{N}(\bm{0}, \mathbf{I})$ through conditional sampling: for any $t\in[T]$, let
\begin{align}
\label{eq:forward diffusion conditional}
    q(\bm{x}_t | \bm{x}_{t-1}) = \mathcal{N}(\bm{x}_t; \sqrt{1 - \beta_t} \bm{x}_{t-1}, \beta_t \mathbf{I}), 
\end{align}
where $\beta_1, \beta_2, \ldots, \beta_T$ are scheduling parameters. 
According to the properties of Gaussian distributions,
Equation \ref{eq:forward diffusion conditional} implies that we can directly sample $\bm{x}_t$ for any $t\in[T]$ based on:
\begin{align}
\label{eq:forward diffusion cumulative}
    q(\bm{x}_t | \bm{x}_0) = \mathcal{N}(\bm{x}_t; \sqrt{\bar{\alpha}_t} \bm{x}_0, (1 - \bar{\alpha}_t) \mathbf{I}),
\end{align}
where $\bar{\alpha}_t = \Pi_{s=1}^t (1 - \beta_s)$.
Based on Equation \ref{eq:forward diffusion cumulative}, Ho et al. \cite{ho2020denoising} proposed to train DDPM by minimizing a regression loss:
\begin{align}
\label{eq: DDPM training loss}
    L_{\mathrm{DDPM}}(\theta) = \mathbb{E}_{\bm{x}_0\sim q_0, \bm{\epsilon} \sim \mathcal{N}(\bm{0}, \mathbf{I}), t\in\mathcal{U}(T)} \big\| \bm{\epsilon} - \bm{\epsilon}_\theta (\bm{x}_t , t)  \big\|_2^2,
\end{align}
where $\bm{x}_t = \sqrt{\bar{\alpha}_t} \bm{x}_0 + \sqrt{1 - \bar{\alpha}_t} \bm{\epsilon}$ denotes the noisy image corresponding to $\bm{x}_0$ at $t$-th forward step, $\bm{\epsilon}_\theta: \mathbb{R}^d \times [0,1] \rightarrow \mathbb{R}^d$ stands for a neural net with weights $\theta$ for noise prediction from $\bm{x}_t$, and $\mathcal{U}(T)$ denotes the uniform distribution over the set $\{1, 2, \ldots, T\}$.
Once the model $\bm{\epsilon}_\theta$ is trained, we can then generate synthetic images from Gaussian noise via the following reverse (or denoising) process: for $t = T, \ldots, 1$,
\begin{align}
\label{eq:DDPM reverse process}
    \bm{x}_{t-1} = \frac{1}{\sqrt{\alpha_t}} \big( \bm{x}_t - \frac{1 - \alpha_t}{\sqrt{1 - \bar{\alpha}_t}} \bm{\epsilon}_{\theta}(\bm{x}_t, t) \big) + \sqrt{\beta_t} \bm{\delta},
\end{align}
where $\alpha_t = 1 - \beta_t$, and $\bm{\delta}\sim\mathcal{N}(\bm{0}, \mathbf{I})$.
For simplicity, denote by $g_\theta(\bm{x}_t, t)$ the right-hand side of Equation \ref{eq:DDPM reverse process}, and $g_{\theta}: \mathbb{R}^d \times [0,1] \rightarrow \mathbb{R}^d$ can be viewed as one-step DDPM denoising. 

\shortsection{Define DDPM Guidance Loss}
Now that we have explained DDPM's forward and backward processes, we can introduce the detailed design of our fine-tuning schemes.
Borrowing the insights from previous sections, our method aims to fine-tune a DDPM model to directly generate the required subset of boundary-adjacent unlabeled data points for more efficient SSAT.
Similar to our selection-based methods, we consider three designs: \emph{prediction confidence-guided generation} (PCG), \emph{latent clustering-guided generation with k-means} (LCG-KM), and \emph{latent clustering-guided generation with Gaussian mixture models} (LCG-GMM).
Specifically, we define a guidance loss $\ell_{\mathrm{guide}}: \mathbb{R}^d \rightarrow \mathbb{R}_{+}$ corresponding to each design, which will serve as a fine-tuning regularization term to bias the diffusion model to generate more boundary-adjacent samples.

\shortersection{PCG}
The first design utilizes the prediction confidence of the intermediate model $f_{\hat{\theta}}$ and defines the guidance loss as
\begin{align}
\label{eq: guidance loss PC}
    \ell_{\mathrm{PC}}(\bm{x}) = \mathrm{Conf}(\bm{x}; f_{\hat{\theta}}).
\end{align}
A larger value of $\ell_{\mathrm{PC}}(\bm{x})$ suggests that $\bm{x}$ is more likely to be closer to the model's decision boundary. 
As we've explained in Section \ref{sec:lcs}, the prediction confidence of a neural network classifier can be an unreliable estimator.
Therefore, we propose alternative guidance losses using more robust latent clustering-based approaches.
Based on different clustering techniques, we again introduce two possible variants.

\shortersection{LCG-KM}
For LCG-KM, we first obtain the latent embeddings of all the labeled data with respect to the intermediate model, namely $\{\bm{z} = h_{\hat{\theta}}(\bm{x}): \bm{x}\in\mathcal{S}_l \}$, and then apply k-means clustering algorithm to partition the embeddings into $k$ clusters. 
Note that this step differs from LCS-KM, as we no longer pre-generate the unlabeled dataset.
Formally, let $\{\bm{\mu}_1, \bm{\mu}_2, \ldots, \bm{\mu}_k\}$ denote the centroids corresponding to the $k$ returned clusters, then we define the guidance loss for LCG-KM as:
\begin{align}
\label{eq: guidance loss KM}
    \ell_{\mathrm{KM}}(\bm{x}) = d_2(\bm{x}; h_{\hat{\theta}}) - d_1(\bm{x}; h_{\hat{\theta}}),
\end{align}
where $d_1(\bm{x}; h_{\hat\theta}) = \min_{j\in[k]} \|h_{\hat{\theta}}(\bm{x}) - \bm{\mu}_j \|_2$ denotes the Euclidean distance between the latent embeddings of $\bm{x}$ and the closest centroid, $d_2(\bm{x}; h_{\hat\theta}) = \min_{j\in[k], j\neq j_1} \|h_{\hat{\theta}}(\bm{x}) - \bm{\mu}_j \|_2$ is the second smallest distance, and $j_1$ stands for the index of the closest centroid.
By definition, we know $\ell_{\mathrm{KM}}(\bm{x})$ is always non-negative, and a smaller value of $\ell_{\mathrm{KM}}(\bm{x})$ indicates that the input $\bm{x}$ is more likely to be a boundary-adjacent point.

\shortersection{LCG-GMM}
Similarly, we can define an informative guidance loss based on the coefficients of the Gaussian mixture models. 
Let $\{\bm\mu_j, \mathbf{\Sigma}_j)$ be the mean and covariance parameters returned by GMM with respect to the latent embeddings $\{\bm{z} = h_{\hat{\theta}}(\bm{x}): \bm{x}\in\mathcal{S}_l \}$.
Let $p_j(\bm{z})$ be the $j$-th GMM posterior probability defined by Equation \ref{eq:GMM posterior probability}.
Then, the guidance loss is defined as:
\begin{align}
\label{eq: guidance loss GMM}
    \ell_{\mathrm{GMM}}(\bm{x}) = \tilde{p}_1(\bm{x}; h_{\hat{\theta}}) - \tilde{p}_2(\bm{x}; h_{\hat{\theta}}),
\end{align}
where $\tilde{p}_1(\bm{x}; h_{\hat\theta}) = \argmax_{j} p_j(h_{\hat\theta}(\bm{x}))$ is the highest GMM posterior probability, and $\tilde{p}_2(\bm{x}; h_{\hat\theta})$ is the second highest one.

\shortsection{Guided DDPM Fine-Tuning}
To utilize our designed guidance losses in DDPM fine-tuning, we propose to first denoise $\bm{x}_t$ via a single DDPM reverse step $g_{\theta}(\bm{x}_t, t)$, where $\bm{x}_t$ is a noisy image at $t$-th time step defined by the forward process, and then apply the respective guidance loss to $g_{\theta}(\bm{x}_t, t)$.
For any sample $\bm{x}_0 \sim q_0$, time $t \sim \mathcal{U}(T)$ and noise $\bm{\epsilon}\sim \mathcal{N}(0, \mathbf{I})$, we define the following loss function $L_{\mathrm{reg}}(\theta)$ that will be used as an additional regularization term to fine-tune DDPM:
\begin{align}
    L_{\mathrm{reg}}(\theta) = \mathbb{E}_{\bm{x}_0, t, \bm{\epsilon}} \big[\ell_{\mathrm{guide}}\big( g_{\theta}(\sqrt{\bar{\alpha}_t} \bm{x}_0 + \sqrt{1 - \bar{\alpha}_t} \bm{\epsilon}, t) \big)\big],
\end{align}
where $\ell_{\mathrm{guide}}$ corresponds to Equation \ref{eq: guidance loss PC} when we use prediction confidence, and 
refers to Equations \ref{eq: guidance loss KM} and \ref{eq: guidance loss GMM} when using LCG-KM and LCG-GMM, respectively.
Putting all the pieces together, the fine-tuning loss for direct generation of boundary-adjacent data is defined as:
\begin{align}
\label{eq: total DDPM training loss}
    L_{\mathrm{tot}}(\theta) = L_{\mathrm{DDPM}}(\theta) +  \lambda \cdot L_{\mathrm{reg}}(\theta),
\end{align}
where $\lambda \geq 0$ balances the trade-off between high-quality image generation captured by the original DDPM loss and boundary-seeking data generation driven by our guidance loss. By penalizing the model weights through the regularization term, our method gradually pushes $\theta$ towards higher probabilities of generating boundary-adjacent data using \emph{stochastic gradient descent} (SGD).
Our later experiments show that when starting from a pre-trained DDPM (i.e., on CIFAR-10), a few fine-tuning epochs are sufficient to guide the diffusion model to generate the desirable subset of synthetic images.

\section{Experiments}
\label{sec:experiment}

In this section, we evaluate the performance of our data selection methods (Section \ref{sec:main results selection}), discuss their computational advantages compared to full SSAT (Section \ref{sec:computational benefits}), and study the effectiveness of our guided DDPM fine-tuning schemes (Section \ref{sec:main results generation}). All our experiments are conducted on a single A100 GPU. Below, we briefly introduce the main experimental setup, where full details are deferred to Appendix \ref{append:experimental details}.

\shortsection{Dataset}  
We consider two benchmark image datasets, SVHN \cite{netzer2011reading} and CIFAR-10 \cite{alex2009learning}, and two scenarios in terms of the source of the unlabeled dataset $\mathcal{S}_u$ for evaluating our methods: (i) \emph{external}--using an external dataset, and (ii) \emph{generative}--using a synthetic dataset generated by some pre-trained generative model. 
For external scenarios, we follow the implementation protocols outlined by Carmon et al. \cite{carmon2019unlabeled}.
To be more specific, SVHN models are trained on $73$K labeled digit images from the original dataset, supplemented by $531$K extra unlabeled SVHN images.  
For CIFAR-10, the original dataset comprises $50$K labeled training images and $10$K labeled test images, where models trained using SSAT are augmented by an external unlabeled dataset $\mathcal{S}_u$ with $500$K images sampled from the $80$M Tiny Images (80M-TI) dataset. 
For the generative setup, we use $1$M synthetic images generated by DDPM for both datasets, following the protocols in Gowal et al. \cite{gowal2021improving}. 

\shortsection{Configuration}
We consider $\ell_\infty$ perturbations with $\epsilon = 0.015$ on SVHN and $\epsilon = 0.031$ on CIFAR-10.
We adopt WideResNet (WRN) architectures and train models using TRADES~\cite{zhang2019theoretically} on both labeled and unlabeled data, incorporating pseudo labels generated by the intermediate model.
In Appendix \ref{app:more_analysis}, we evaluate the generalizability of our methods for varying $\ell_\infty$ perturbation size, vanilla PGD-based adversarial training with varying attack steps, and $\ell_2$ perturbations.
To standardize training, an epoch is defined as processing $50$K data points, regardless of the total dataset size. 
Prior works also adopted this approach~\cite{carmon2019unlabeled, gowal2021improving}, which ensures that training time per epoch remains consistent across datasets of varying sizes. 

\begin{table*}[!t]
\centering
\caption{SSAT performance under various unlabeled data selection schemes. We consider $\ell_\infty$ perturbations with $\epsilon = 0.015$ for SVHN, and $\ell_\infty$ perturbations with $\epsilon = 0.031$ for CIFAR-10. For each configuration, we underline the best robust accuracies.}
\label{table:SVHN+CIFAR10}
\vspace{-0.05in}
\resizebox{0.96\textwidth}{!}{
\begin{tabular}{l l l | c c c | c | c c c | c}
    \toprule
    \multirow{2.4}{*}{\textbf{Dataset}} & \multirow{2.4}{*}{$\bm{\alpha}$} & \multirow{2.4}{*}{\textbf{Method}} & \multicolumn{4}{c|}{\textbf{External Unlabeled}} & \multicolumn{4}{c}{\textbf{DDPM-Generated}} \\
    \cmidrule{4-11}
    & & & \small{\textbf{Clean} ($\%)$} & \small{\textbf{PGD} ($\%)$} & \small{\textbf{AA} ($\%)$} &  \small{\textbf{Time} (h)} & \small{\textbf{Clean} ($\%)$} & \small{\textbf{PGD} ($\%)$} & \small{\textbf{AA} ($\%)$} & \small{\textbf{Time} (h)} \\
    \midrule
    \multirow{13.5}{*}{SVHN} & $0\%$ & No Sel. & $94.4 $  & $75.6$ & $68.3$ & $2.62$ & $94.4$ & $75.6$ & $68.3$ & $2.62$  \\ 
    & $100\%$ & No Sel. & $97.1$  & $86.0$ & $75.3$  & $8.38$ & $97.4$  & $86.3$ & $75.2$ & $17.41$ \\ 
    \cmidrule{2-11}
    & \multirow{4}{*}{$1\%$} 
    & Random    & $94.2$  & $78.4$ & $70.3 $ & $3.28$  & $95.1$  & $80.3$ & $70.5$ & $3.28$   \\ 
    & & PCS  & $95.4$  & $81.8$ & $71.6$ & $3.75$ & $95.4$  & $81.5$ & $71.5$ & $4.08$ \\ 
    & & LCS-GMM   & $95.1$  & $81.8$ & $72.1$ & $3.97$ & $95.6$  & $81.2$ & $71.8$ & $4.27$ \\ 
    & & LCS-KM   & $95.2$  & $\underline{82.7}$ & $\underline{72.9}$ & $3.90$ & $96.4$  & $\underline{82.4}$ & $\underline{72.6}$ & $4.18$ \\ 
    \cmidrule{2-11}
    & \multirow{4}{*}{$10\%$} 
    & Random    & $95.3$  & $82.0$ & $73.0$ &  $3.28$ & $95.6$  & $83.6$ & $72.4$ & $3.28$   \\ 
    & & PCS  & $96.1$ & $82.8$ & $74.2$ & $3.85$ & $95.3$  & $84.0$ & $73.5$ & $4.08$  \\ 
    & & LCS-GMM   & $96.2$  & $83.0$ & $74.3$ & $3.97$ & $95.5$  & $84.1$ & $74.1$ & $4.27$ \\ 
    & & LCS-KM   & $96.1$  & $\underline{86.3}$ & $\underline{75.2}$ & $3.90$ & $96.6$  & $\underline{86.6}$ & $\underline{74.8}$ & $4.18$ \\ 
    \cmidrule{2-11}
    & \multirow{4}{*}{$20\%$}
    & Random    & $96.2$  & $82.3$ & $73.2$ & $3.28$ & $96.3$  & $84.6$ & $74.0$ & $3.28$ \\ 
    & & PCS  & $96.3$  & $83.9$ & $74.7$ & $3.75$ & $96.6$  & $85.4$ & $74.5$ & $4.08$ \\ 
    & & LCS-GMM   & $96.4$  & $84.3$ & $75.0$ & $3.97$ & $95.8$  & $85.5$ & $74.7$ & $4.27$ \\ 
    & & LCS-KM   & $96.2$  & $\underline{86.6}$ & $\underline{75.1}$ & $3.90$ & $96.7$  & $\underline{87.2}$ & $\underline{75.3}$ & $4.18$ \\ 
    \midrule[0.8pt]
    \multirow{13.5}{*}{CIFAR-10} & $0\%$ & No Sel. & $84.9$  & $55.4$ &$49.2$ & $10.69$  & $84.9$  & $55.4$ &$49.2$ & $10.69$ \\ 
    & $100\%$ & No Sel. & $89.7$ & $62.5$ & $58.6$  & $28.50$ & $85.7$  & $61.8$ & $58.4$ & $57.13$ \\
    \cmidrule{2-11}
    & \multirow{4}{*}{$1\%$} 
    & Random    & $83.2$  & $54.5$ &$49.8$ & $14.25$ & $83.3$ & $54.3$ & $49.6$ & $14.25$   \\ 
    & & PCS  & $84.9$  & $55.4$ & $51.5$ & $14.73$ & $84.3$  & $55.2$ &$50.7$ & $15.16$  \\ 
    & & LCS-GMM   & $84.2$  & $55.1$ & $52.1$ & $14.89$ & $85.8$  & $54.6$ & $51.8$ & $15.36$ \\ 
    & & LCS-KM   & $85.6$  & $\underline{56.4}$ & $\underline{52.9}$ & $14.78$ & $85.9$  & $\underline{56.1}$ & $\underline{52.5}$ & $15.27$  \\ 
    \cmidrule{2-11}
    &\multirow{4}{*}{$10\%$} 
    & Random    & $85.2$  & $56.0$ & $52.4$ & $14.25$  & $85.0$  & $56.5$ & $52.6$ & $14.25$  \\ 
    & & PCS  & $85.3$ & $56.9$ & $53.7$ & $14.73$ & $85.4$  & $56.9$ & $54.1$ & $15.16$   \\ 
    & & LCS-GMM   & $85.9$  & $57.1$ & $54.2$ & $14.89$ & $85.8$  & $57.1$ & $54.5$ & $15.36$ \\ 
    & & LCS-KM   & $87.2$  & $\underline{58.2}$ & $\underline{55.3}$ & $14.78$ & $86.1$  & $\underline{58.0}$ & $\underline{55.8}$ & $15.27$ \\ 
    \cmidrule{2-11}
    & \multirow{4}{*}{$20\%$} 
    & Random    & $87.1$  & $57.5$ & $54.2$ & $14.25$ & $85.4$  & $57.2$ &$54.2$ & $14.25$  \\ 
    & & PCS  & $87.0$  & $57.9$ &$54.5$ & $14.73$ & $85.6$  & $58.0$ &$55.0$ & $15.16$  \\ 
    & & LCS-GMM   & $87.0$  & $58.2$ &$55.5$ & $14.89$ & $85.9$  & $58.6$ &$55.8$ & $15.36$ \\ 
    & & LCS-KM   & $88.7$  & $\underline{60.7}$ & $\underline{57.8}$ & $14.78$ & $85.5$  & $\underline{60.2}$ & $\underline{57.2}$ & $15.27$ \\
    \bottomrule
\end{tabular}
}
\end{table*}

\shortersection{SSAT} With external data, we train SSAT algorithms for a total of $200$ epochs using stepwise learning rate scheduling, whereas training is extended to $400$ epochs under generative setups. 
To ensure fair comparisons, models are saved every $25$ epoch, and the model achieving the highest robust accuracy is selected as the \emph{best} model. This approach aligns with the early stopping practices widely used in prior literature \cite{zhang2019theoretically,rice2020overfitting}. When the ratio of selected unlabeled data $\alpha \in \{10\%, 20\%\}$, we consistently observe peak robust accuracy around $100$ epochs for CIFAR-10 and $75$ epochs for SVHN. Conversely, when no data selection is applied, peak robustness is delayed to approximately $200$ epochs. The total training time is reported as the duration required to achieve optimal performance.

\shortersection{Intermediate Model} 
Our data selection and guided diffusion approaches require training an intermediate model $f_{\hat{\theta}}$, which adopts the same architecture as the final model. The intermediate model fulfills two critical roles: it not only facilitates the pseudo-labeling of unlabeled data but also constitutes a necessary component for implementing any of our data reduction strategies. To be more specific, we train the intermediate model $f_{\hat{\theta}}$ using standard supervised learning for $100$ epochs on WRN-28-10, which takes approximately $55$ minutes. We exclude this time from the reported total runtime, as this step is integral to the overall pipeline of any tested SSAT algorithm. 

\shortsection{Evaluation Metric}
To evaluate the robustness of the final model returned by SSAT, we mainly employ multi-step PGD attacks (PGD)~\cite{madry2017towards}, due to their broad adoption in adversarial ML. 
We also report robust accuracies against AutoAttack (AA)~\cite{croce2020reliable} to provide a more rigorous robustness evaluation, as well as the model's clean accuracy on normal inputs (Clean).

\subsection{Strategic Unlabeled Data Selection}
\label{sec:main results selection}

Table \ref{table:SVHN+CIFAR10} documents our evaluation results on SVHN and CIFAR-10, showing the efficacy of our data reduction schemes when incorporated into semi-supervised adversarial training under varying data ratios $\alpha\in\{1\%, 10\%, 20\%\}$, compared with baseline TRADES ($\alpha = 0\%$) and full SSAT ($\alpha=100\%$). 


\begin{figure*}[!t] 
    \centering
    \subfloat[Vanilla AT]{
        \includegraphics[width=0.235\linewidth]{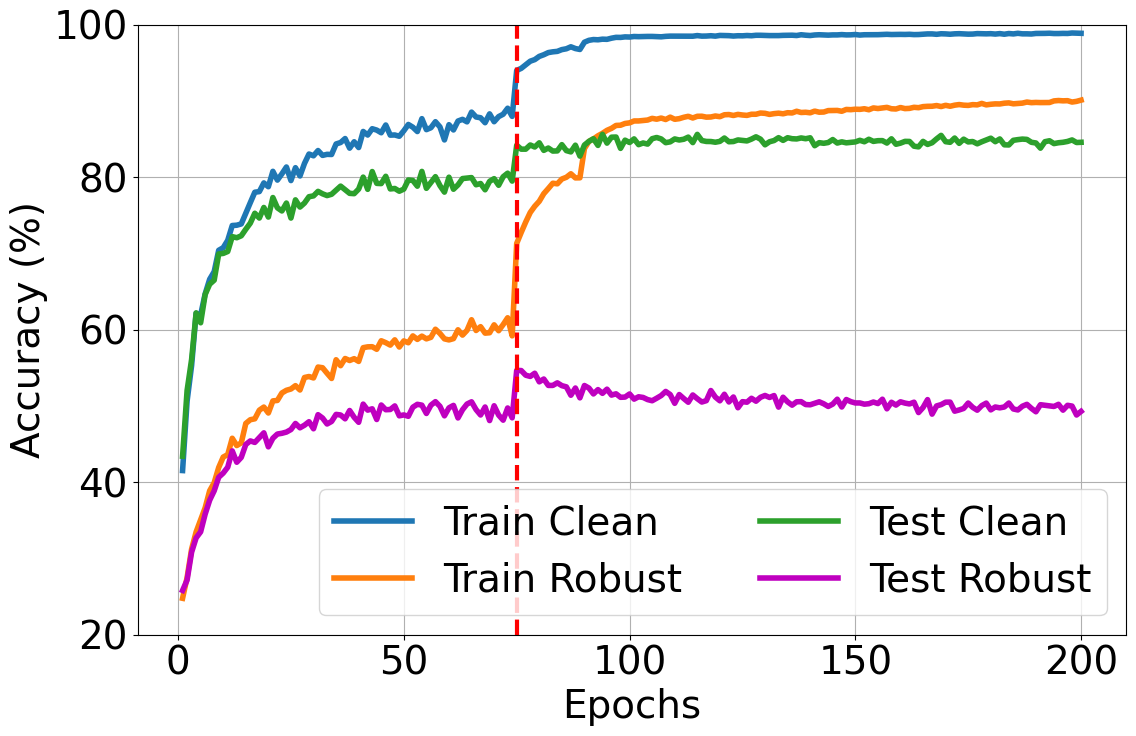}
        \label{fig:0per}
    }
    \subfloat[Random Selection ($\alpha=10\%$)]{
        \includegraphics[width=0.235\linewidth]{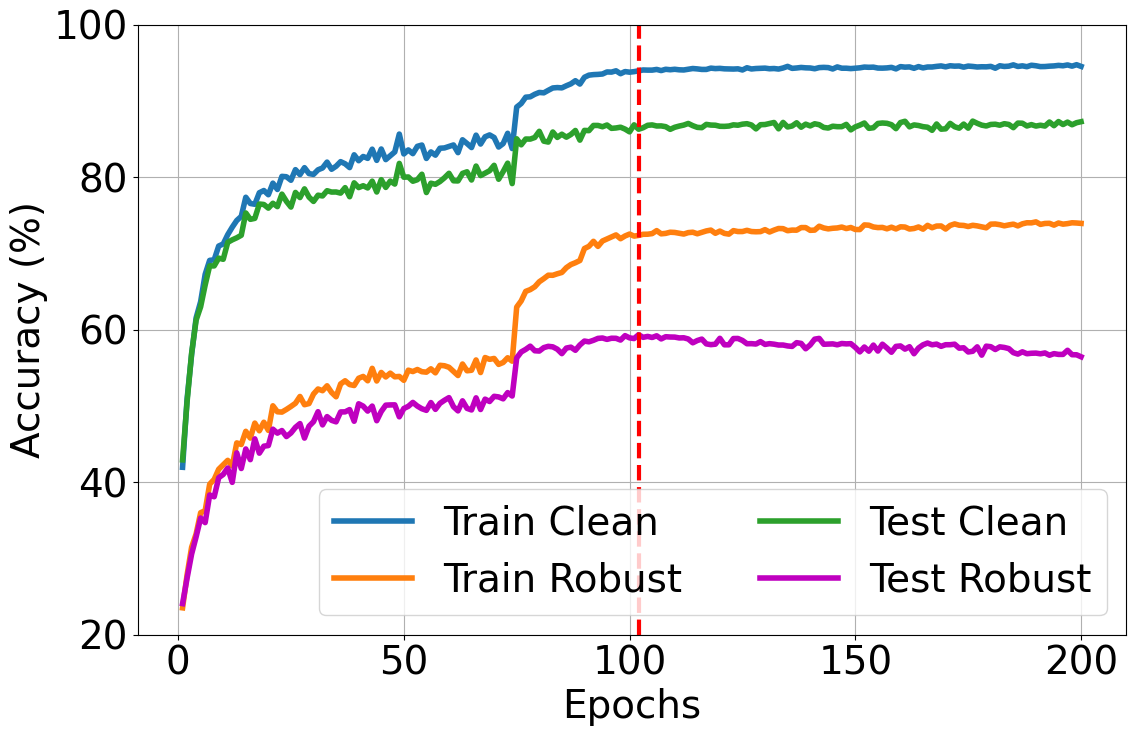}
        \label{fig:10percrandom}
    }
    \subfloat[LCS-KM ($\alpha=10\%$)]{
        \includegraphics[width=0.235\linewidth]{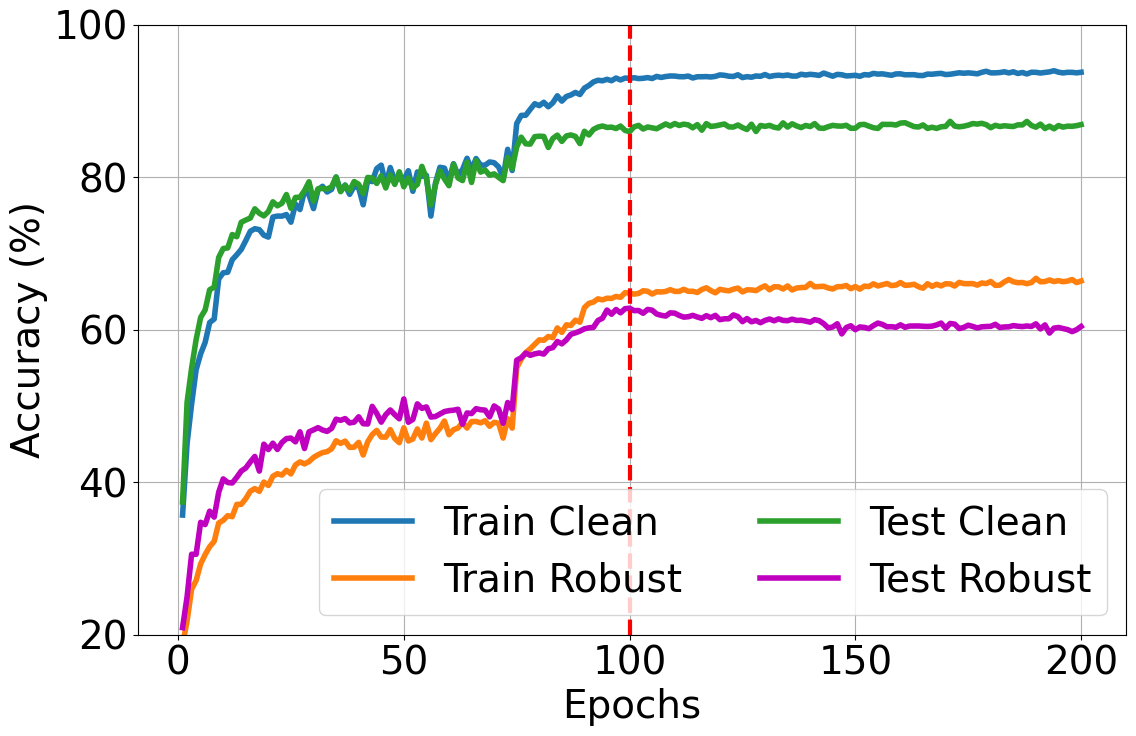}
        \label{fig:10perc}
    }
    \subfloat[Full Utilization ($\alpha=100\%$)]{
        \includegraphics[width=0.235\linewidth]{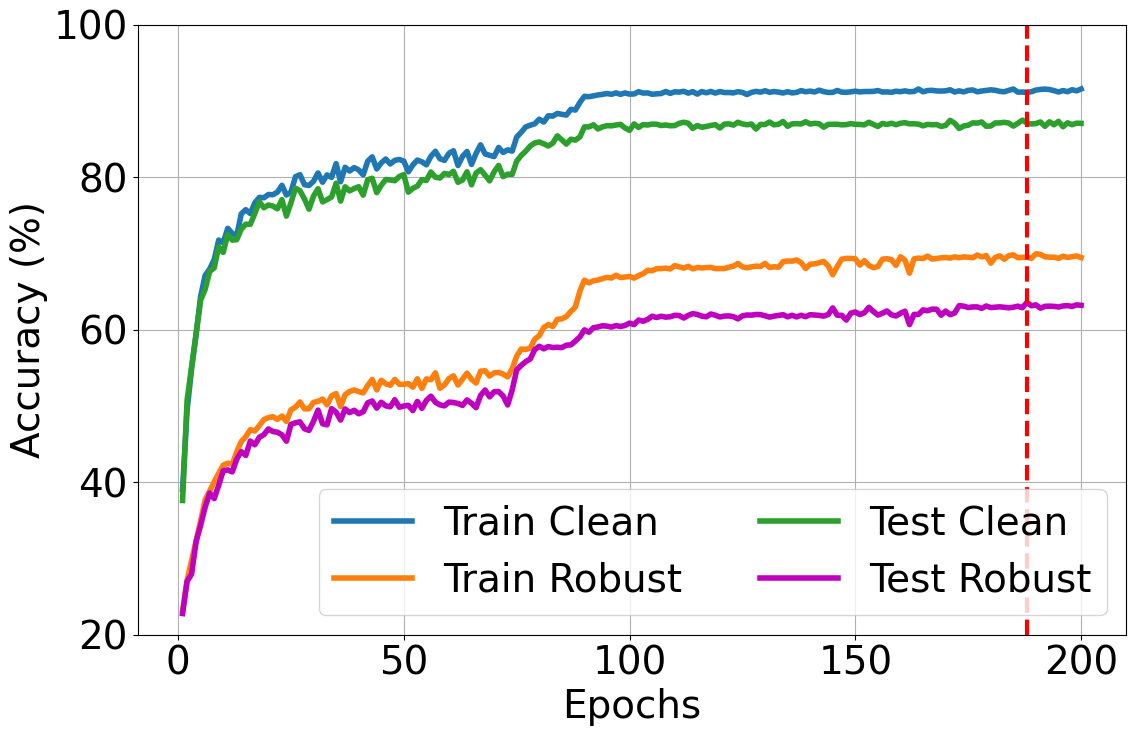}
        \label{fig:100perc}
    }
    \vspace{-0.1in}
    \caption{Illustration of standard and robust accuracy curves of SSAT on CIFAR-10 with different configurations of unlabeled data selection from the external $500$K Tiny Images: (a) No extra data, (b) random selection with $\alpha = 10\%$, (c) LCS-KM with $\alpha = 10\%$, and (d) utilizing all $500$K unlabeled data.}
    \vspace{-0.1in}
    \label{fig:convergence}
\end{figure*}

\shortsection{SVHN} Specifically for SVHN, we adopt a WRN-16-8 architecture. Employing our LCS-KM strategy with $10\%$ or $20\%$ of extra unlabeled data achieves PGD robust accuracies of $86.3\%$ and $86.6\%$, respectively, which matches and even slightly outperforms the $86.0\%$ robustness achieved using the entire unlabeled dataset.
In contrast, random selection with $20\%$ of the data yields a robust accuracy of only $82.3\%$, underscoring the superior efficacy of informed selection strategies such as LCS-KM. 
The robust accuracies evaluated using AutoAttack demonstrate similar patterns compared to those using multi-step PGD Attacks.
In addition, alternative methods, including PCS and LCS-GMM, consistently outperform random selection, confirming the effectiveness of our insights. 
Notably, the robustness of models trained using SSAT peaks when the boundary ratio parameter $\beta$ is inversely proportional to the selection ratio $\alpha$, suggesting that smaller $\beta$ values are favorable as the data inclusion proportion increases. 
We provide detailed sensitivity analyses of these hyperparameters in Section \ref{sec:hyperparameter tuning}.

Our evaluations with $1$M DDPM-generated data highlight that when using $10\%$ to $20\%$ of unlabeled data selected via LCS-KM, SSAT can achieve comparable robustness to the SSAT baseline.
Since all SSAT algorithms listed under the column ``DDPM-Generated'' in Table \ref{table:SVHN+CIFAR10} require the same amount of pre-generation time, we exclude it from the total runtime.
In particular, training SSAT with reduced data requires only $75$ epochs to achieve the peak robust accuracy, while training with the full dataset requires up to $400$ epochs. 
Such a reduction in computation time underscores the practical advantages of the proposed data selection strategies, which are further elaborated in Section \ref{sec:computational benefits}.
Besides, we observe in our experiments that using ground-truth labels reveals negligible differences compared to pseudo-labeled data, indicating that the primary performance gains stem from the selection strategy itself rather than label accuracy (see Table \ref{tab:SVHN-Result_truelabels} in Appendix \ref{append:additional experiments}).

\shortsection{CIFAR-10}
We adopt a WRN-28-10 architecture for CIFAR-10 experiments.
Across varying ratios $\alpha$ and data selection schemes, Table \ref{table:SVHN+CIFAR10} reveals that selecting $20\%$ of the external data using LCS-KM attains PGD robust accuracy of $60.7\%$, close to the $62.5\%$ obtained with the entire dataset. In contrast, random selection with the same ratio achieves only $57.5\%$, underscoring the advantages of our strategic LCS technique.
Similar to SVHN, we extend our analysis to CIFAR-10 with synthetically generated data. Using $1$M DDPM-generated CIFAR-10-like images, we discover that using just $20\%$ of the generated data via LCS-KM results in performance nearly identical to that achieved with the full dataset. The above observations are also consistent across other settings, such as $\ell_2$ perturbations and a ResNet-18 model architecture (see Figure \ref{fig:two_graphs_ab} and Table \ref{table:CIFAR10-External-resnet} in the appendix for supporting evidence).

The above findings emphasize the adaptability of our data selection schemes, particularly for LCS-KM, under varying experimental conditions.
When leveraging a small subset of selected external unlabeled data, optimal robust performance was achieved at $100$ epochs of training. In contrast, using the full external or generated datasets without selection requires extended training of $200$ and $400$ epochs, respectively, to reach peak performance. These results validate the efficacy of our latent clustering-based selection approach in enhancing semi-supervised adversarial training. The proposed method not only maintains the robust accuracy of the final model, $\theta_{\mathrm{final}}$, but also effectively accelerates the training process, underscoring its practical utility and scalability in SSAT frameworks. 

\subsection{Computational Benefit of Data Reduction}
\label{sec:computational benefits}

As suggested by the runtime comparisons demonstrated in Table \ref{table:SVHN+CIFAR10}, utilizing smaller, carefully selected unlabeled data can effectively reduce the overall training time of SSAT while maintaining competitive robustness. 
We further visualize the learning curves of SSAT and discuss the computational benefits associated with our proposed methods.
Figure \ref{fig:convergence} depicts the learning curves of vanilla AT and SSAT algorithms on CIFAR-10 with varying ratios of unlabeled data selected from TinyImages. In our setup, one training epoch corresponds to processing $50$K data points, with $10$ PGD steps for training and $20$ for testing. 
When no additional unlabeled data is utilized, the highest test robust accuracy is achieved after approximately $75$ epochs (Figure \ref{fig:0per}). 
Incorporating $10\%$ external unlabeled data, whether through random selection or curated approaches, extends the optimal convergence point to approximately $100$ epochs (Figures \ref{fig:10percrandom} and \ref{fig:10perc}). In contrast, leveraging $100\%$ of the unlabeled data delays the achievement of peak accuracy to around $185$ epochs (Figure \ref{fig:100perc}). These observations suggest that including extra unlabeled data in SSAT increases computational demands.
Our selection schemes reduce redundant information in the dataset, allowing SSAT to achieve high robustness performance with fewer training epochs.

\begin{table}[!t]
\centering
\caption{Performance of full SSAT ($\alpha = 100\%$) with early stopping at epoch $100$ on CIFAR-10 under $\ell_\infty$ perturbations with $\epsilon=0.031$.}
\label{tab:early_stop_cifar10}
\vspace{-0.05in}
\small
\resizebox{0.5\textwidth}{!}{
\begin{tabular}{l l | c c c}
    \toprule
    \textbf{Data Source} & \textbf{Schedule} & \textbf{Clean} ($\%)$ & \textbf{PGD} ($\%)$ & \textbf{AA} ($\%)$ \\
    \midrule
    \multirow{2}{*}{External (500K)} 
    & Cosine & $84.1$ & $54.0$ & $50.4$ \\
    & Stepwise   & $88.0$ & $60.6$ & $57.6$ \\
    \midrule
    \multirow{2}{*}{DDPM (1M)} 
    & Cosine & $78.6$ & $52.1$ & $48.6$ \\
    & Stepwise   & $83.0$ & $58.5$ & $55.9$ \\
    \bottomrule
\end{tabular}
}
\vspace{-0.1in}
\end{table}

\begin{table*}[t]
\centering
\caption{Runtime comparison of SSAT with synthetically generated unlabeled data across various settings on CIFAR-10.}
\vspace{-0.05in}
\small
\label{tab:pipeline_time}
\resizebox{0.95\textwidth}{!}{
\begin{tabular}{ll | ccccc | c}
\toprule
$\bm{\alpha}$ & \textbf{Method} & \textbf{Fine-Tuning} (h) & \textbf{Generation} (h) & \textbf{Selection} (h) & \textbf{\#Epochs} & \textbf{Total Time} (h) & \textbf{PGD} (\%) \\
\midrule
$0\%$ & No Gen. & -- & -- & -- & $100$ & $10.7$ & $55.4$ \\
$100\%$ & Std. DDPM & -- & $3.90$ & -- & $400$ & $61.0$ & $61.8$ \\
$20\%$ & LCS-KM & -- & $3.90$ & $0.97$ & $100$ & $19.1$ & $60.3$ \\
$20\%$ & LCG-KM & $0.43$ & $0.77$ & -- & $100$ & $15.7$ & $60.2$ \\
\bottomrule
\end{tabular}
}
\vspace{-0.05in}
\end{table*}

\begin{table}[t]
\centering
\small
\caption{Performance of our DDPM fine-tuning strategies on CIFAR-10 under varying $\alpha$. The best robust accuracies are underlined.}
\vspace{-0.05in}
\label{tab:cifar10_finetuning}
\resizebox{0.48\textwidth}{!}{
\begin{tabular}{l l |ccc}
\toprule
$\bm\alpha$ & \textbf{Method} & \textbf{Clean} ($\%$) & \textbf{PGD} ($\%$) & \textbf{AA} ($\%$) \\
\midrule
$0\%$   & No Gen. & $84.9$ & $55.4$ & $49.2$ \\
$100\%$ & Std. DDPM  & $85.7$ & $61.8$ & $58.4$  \\
\midrule
\multirow{3}{*}{$1\%$} & PCG & $84.4$ & $55.4$ & $52.8$ \\
    & LCG-GMM & $85.3$ & $55.8$ & $52.6$ \\
    & LCG-KM & $85.2$ & \underline{$56.5$} & \underline{$53.1$} \\
\midrule
\multirow{3}{*}{$10\%$} & PCG & $84.7$ & $56.7$ & $54.2$ \\
    & LCG-GMM & $85.8$ & $57.6$ & $55.0$ \\
    & LCG-KM & $84.2$ & \underline{$57.9$} & \underline{$55.6$} \\
\midrule
\multirow{3}{*}{$20\%$} & PCG & $85.9$ & $58.3$ & $55.9$ \\
    & LCG-GMM & $84.7$ & $58.3$ & $56.1$ \\
    & LCG-KM & $85.7$ & \underline{$60.3$} & \underline{$57.4$} \\
\bottomrule
\end{tabular}
}
\vspace{-0.05in}
\end{table}

Generally, the size of the utilized unlabeled dataset is critical in determining the time required for SSAT to achieve peak robust accuracy. Larger unlabeled datasets usually extend the convergence times, whereas smaller datasets facilitate faster convergence, allowing the model to reach optimal accuracy earlier in training. 
Early stopping offers a computational advantage in scenarios with limited data by preventing overfitting with a reduced number of training epochs. Nevertheless, early stopping proves less effective in full SSAT when working with large-scale unlabeled datasets, as longer training is necessary to achieve peak performance.  
Table \ref{tab:early_stop_cifar10} summarizes the model performance on CIFAR-10 with respect to full SSAT when it is early stopped at the same training epoch as our methods. We tested two learning rate schedulers: \textit{cosine} annealing and \textit{stepwise} decay. The results show that the training process of full SSAT algorithms has not converged at the early-stopped training epoch. 
Importantly, even when comparing at the same number of training epochs, our LCS-KM method with $20\%$ DDPM-generated data achieves $60.2\%$ PGD robust accuracy at epoch $100$, outperforming early-stopped full SSAT by $1.7\%$. This demonstrates that our performance gain stems not only from reduced epochs but also from accelerated convergence enabled by strategically selecting more informative samples.
Due to space limits, full results and detailed discussions are provided in Appendix \ref{app:full SSAT early stop}.
These findings highlight the improved trade-offs of our methods in terms of computational costs, data efficiency, and robustness, offering a better solution to address the challenges of SSAT with extensive datasets.  

\begin{figure*}[t] 
    \centering
    \subfloat[Random Selection ($\alpha = 10\%$)]{
        \includegraphics[height=0.15\linewidth, width=0.235\linewidth]{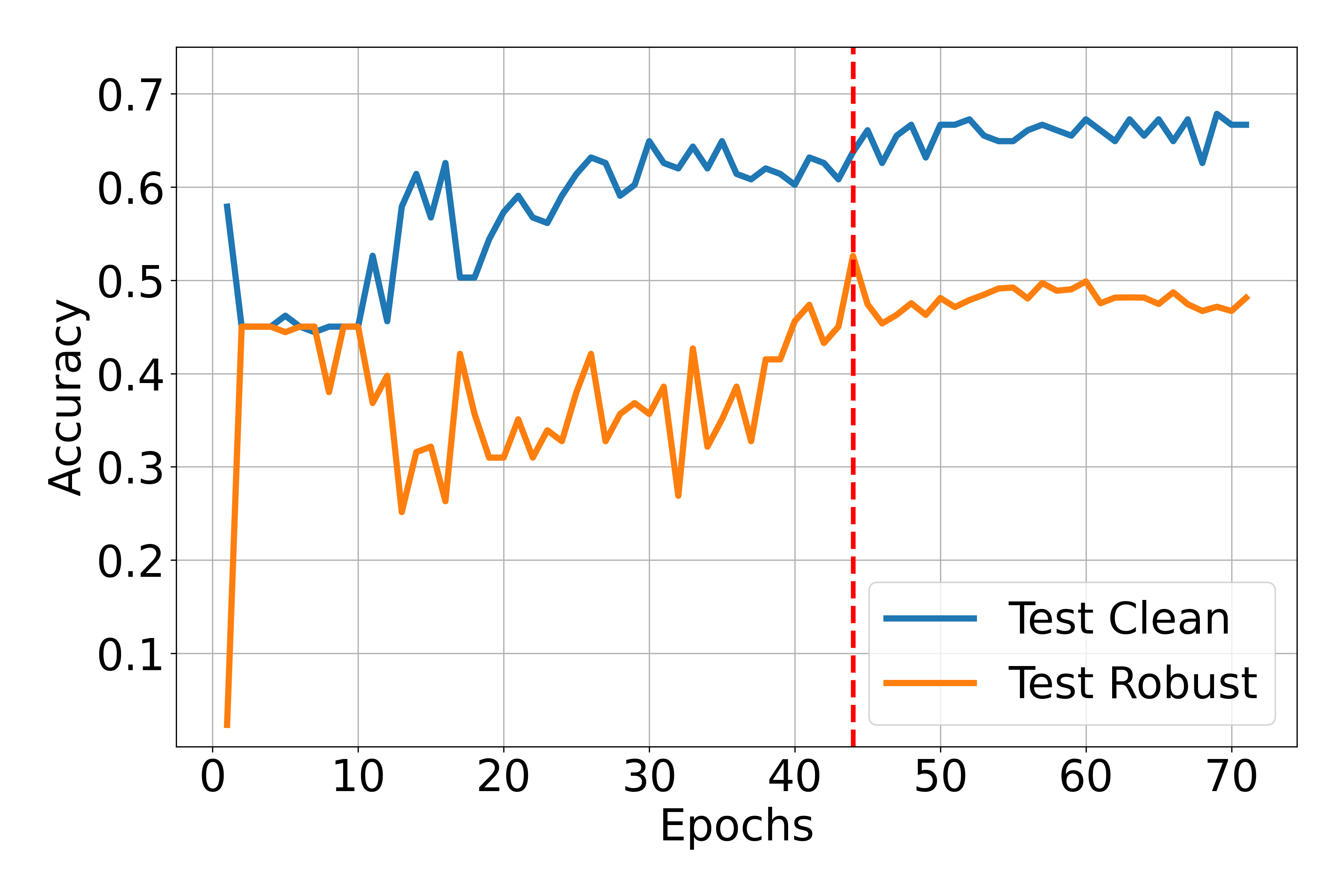}
        \label{fig:10perrandom}
    }
    \subfloat[LCS-KM ($\alpha = 10\%$)]{
        \includegraphics[height=0.15\linewidth, width=0.235\linewidth]{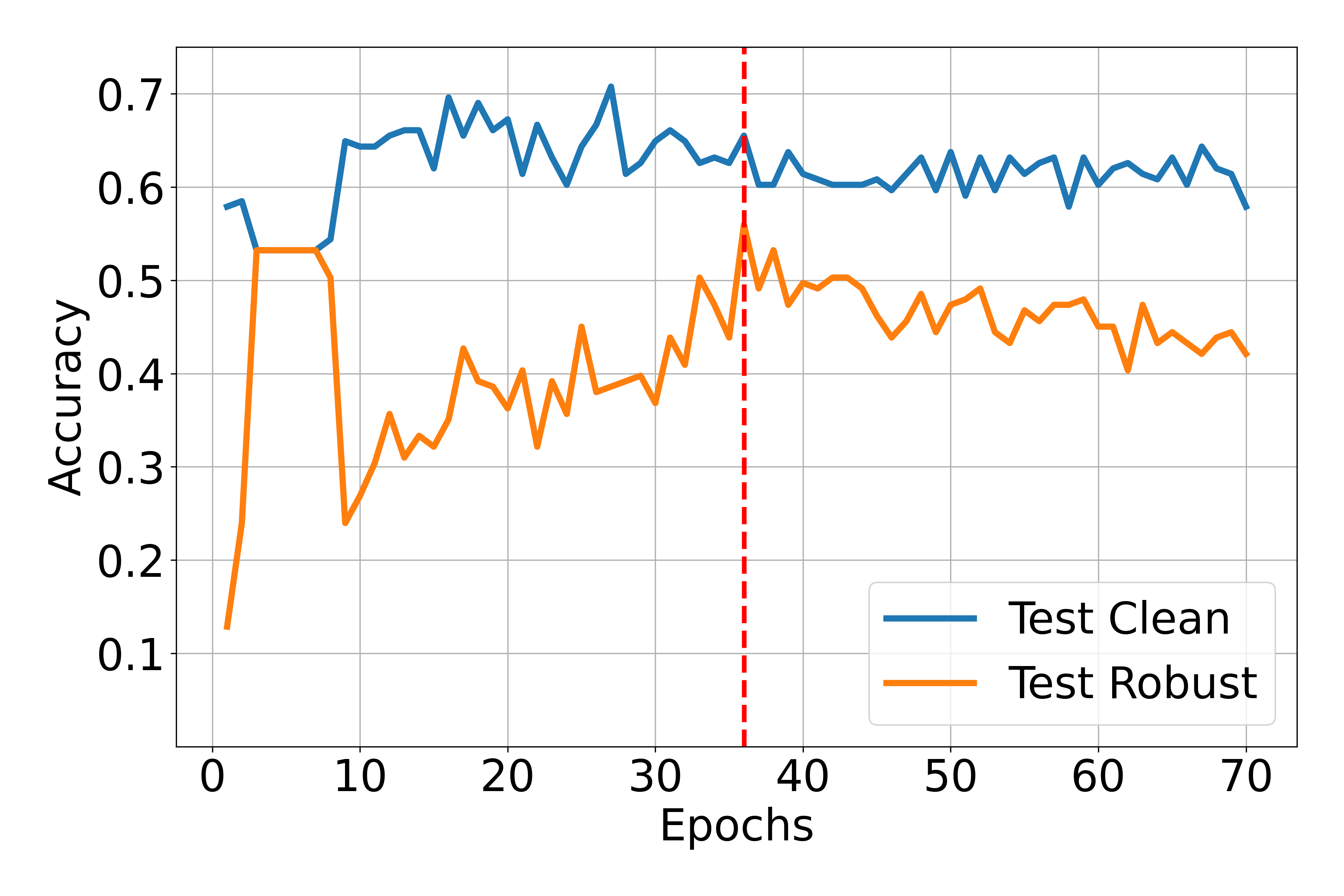}
        \label{fig:10per}
    }
    \subfloat[LCS-KM ($\alpha = 20\%$)]{
        \includegraphics[height=0.15\linewidth, width=0.235\linewidth]{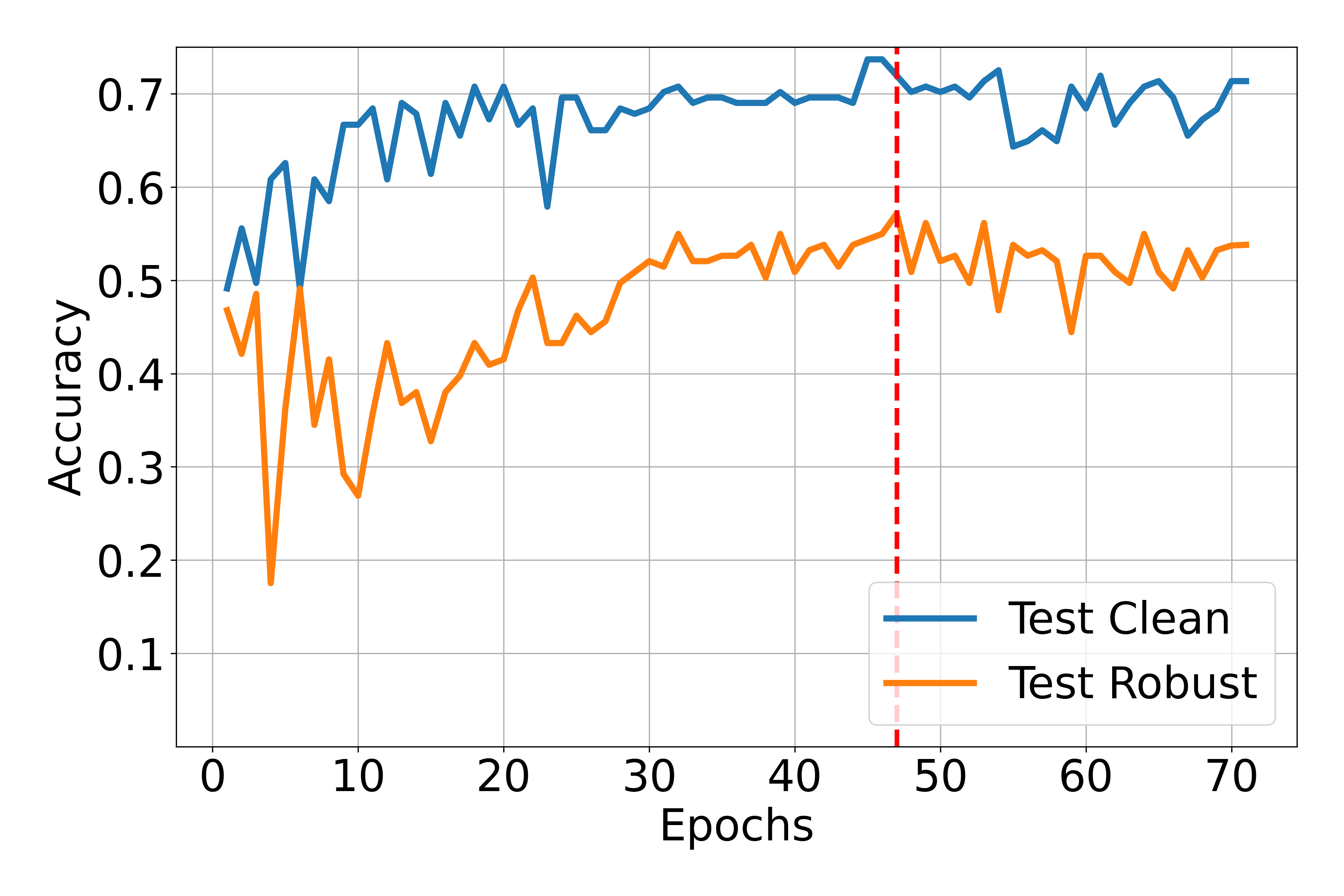}
        \label{fig:20per}
    }
    \subfloat[Full Utilization ($\alpha=100\%$)]{
        \includegraphics[height=0.15\linewidth, width=0.235\linewidth]{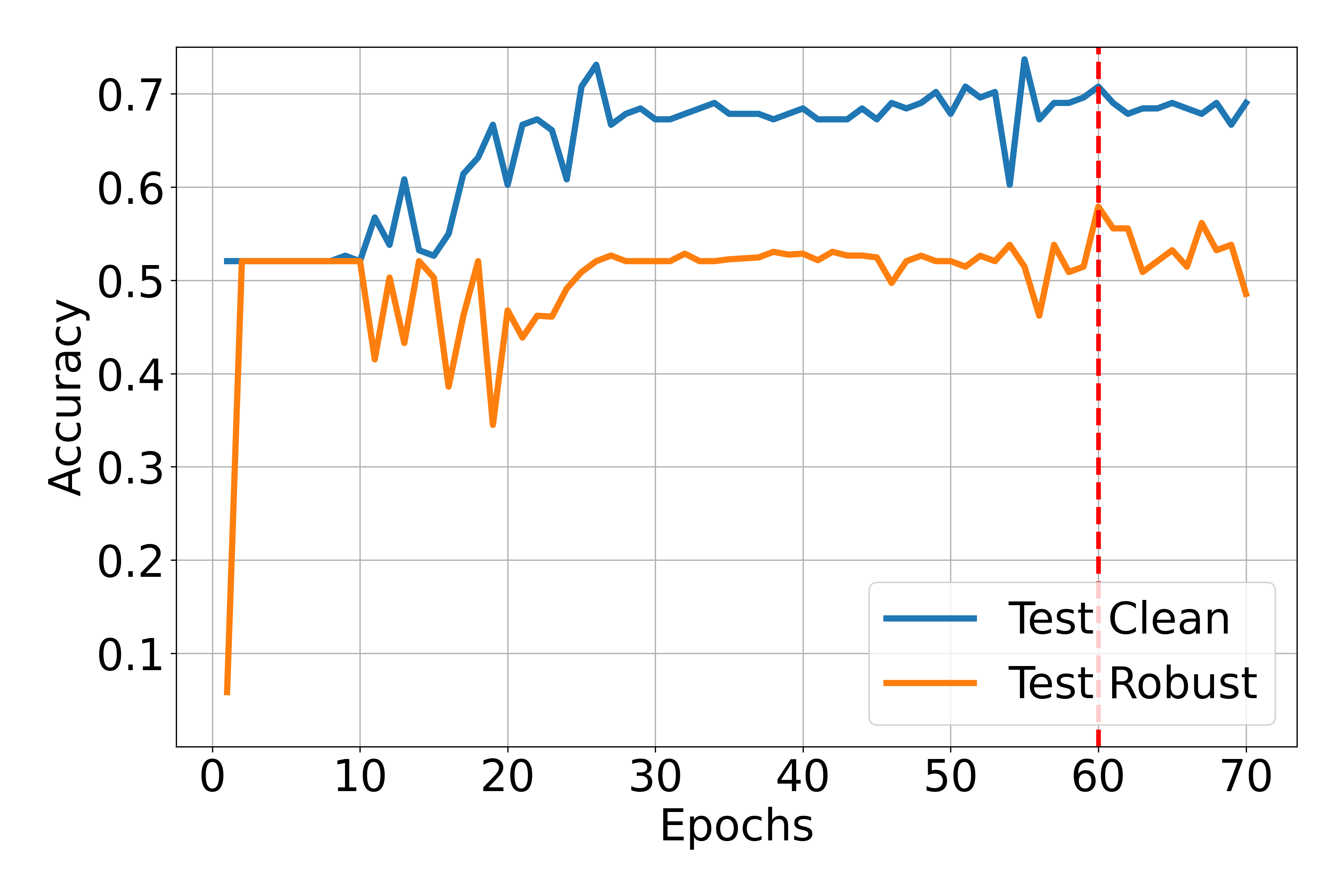}
        \label{fig:100per}
    }
    \vspace{-0.12in}    
    \caption{Standard and robust accuracy curves of SSAT with labeled data from COVIDGRand unlabeled data from CoronaHack, with different selection schemes and ratios: (a) random selection with $\alpha = 10\%$, (b) LCS-KM with $\alpha = 10\%$, (c) LCS-KM with $\alpha = 20\%$, and (d) all unlabeled data $\alpha = 100\%$.}
    \label{fig:four_graphs}
    \vspace{-0.1in}
\end{figure*}

\subsection{Guided DDPM Fine-Tuning}
\label{sec:main results generation}

We evaluate the performance of our guided DDPM fine-tuning methods on CIFAR-10.
Unlike our selection-based schemes, which require pre-generating the entire dataset, here we generate the same amount of unlabeled data, matching the ratio $\alpha$. 
The data-mixing hyperparameters, namely $\alpha$ and $\beta$, are set consistently with those used in strategic selection.
We employ a pretrained DDPM model trained on CIFAR-10 for these guided fine-tuning experiments.  
Table \ref{tab:cifar10_finetuning} presents the results across different $\alpha$ ratios, where our DDPM fine-tuning schemes achieve competitive clean accuracy and robustness performance, matching their selection counterpart. 
For instance, under the ratio scenario $\alpha = 20\%$, leveraging LCG-KM to generate boundary-adjacent data achieves a PGD robust accuracy of $60.3\%$ when integrated in the SSAT framework, which matches the result of $60.2\%$ obtained by LCS-KM. 
Again, latent clustering with k-means consistently achieves the best robustness compared to the other alternatives.
These findings mirror our findings from strategic selection, reinforcing that a small but informative generated set of samples can yield performance comparable to a much larger dataset.

Moreover, we conduct a detailed runtime analysis across different data reduction schemes under the generative setup. 
In particular, we choose to present the best-performing methods with the optimal hyperparameters, namely LCS-KM and LCG-KM under the scenario of $\alpha=20\%$, and compare the end-to-end training times.
We also report the baseline performance of SSAT, where the whole generated dataset ($\alpha = 100\%$) and no generated data ($\alpha = 0\%$) are used, respectively. 
The comparison results are summarized in Table \ref{tab:pipeline_time}.
For clarity, we separately report the computational time of DDPM fine-tuning, data generation, data selection, and the total SSAT training process.
Our DDPM generator's throughput is roughly $71$ images per second, so the approximate data generation times are $200$K images in $46$ minutes, $1$M images in $3.9$ hours, and $100$M images in $16.2$ days.
As a result, our guided DDPM finetuning (LCG-KM) approach reduces the total runtime from $19.1$ hours to $15.7$ hours when compared to LCS-KM.
The improvement will be more pronounced when the total amount of data generated by DDPM is much larger. 
These results highlight the trade-off between robustness gains and computational cost. While generating millions of synthetic samples can enhance robustness, it is computationally expensive. Our guided fine-tuning approach offers a more resource-efficient alternative, reducing SSAT total runtime from $61.0$ hours to $15.7$ hours while preserving strong robustness.

\begin{figure*}[t]
\vspace{-0.05in}
    \centering
    \subfloat[PCS]{
        \includegraphics[width=.31\linewidth]{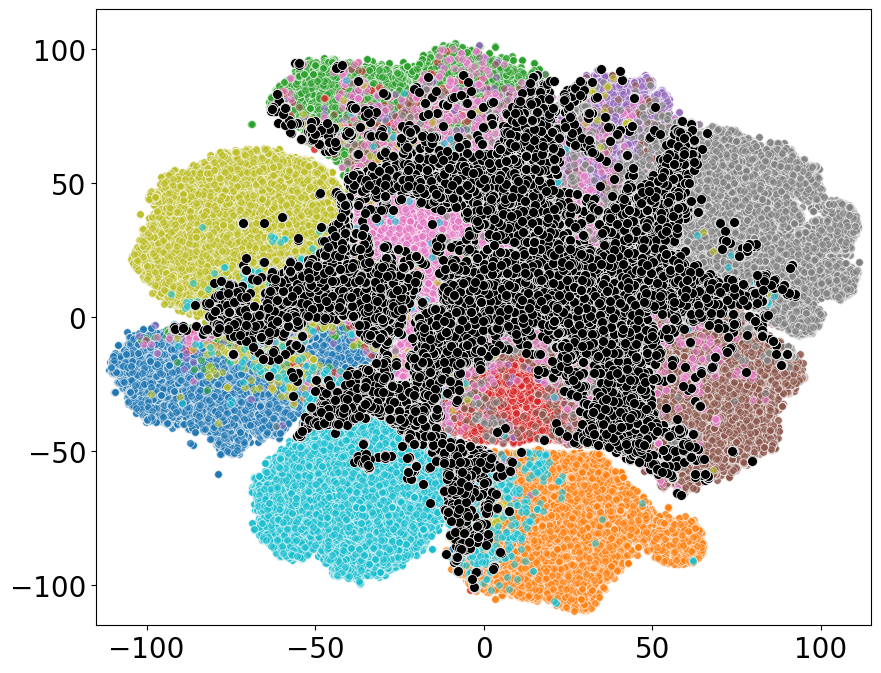}
        \label{fig:PCS_selection}
    }\hfill    
    \subfloat[LCS-GMM]{
        \includegraphics[width=.31\linewidth]{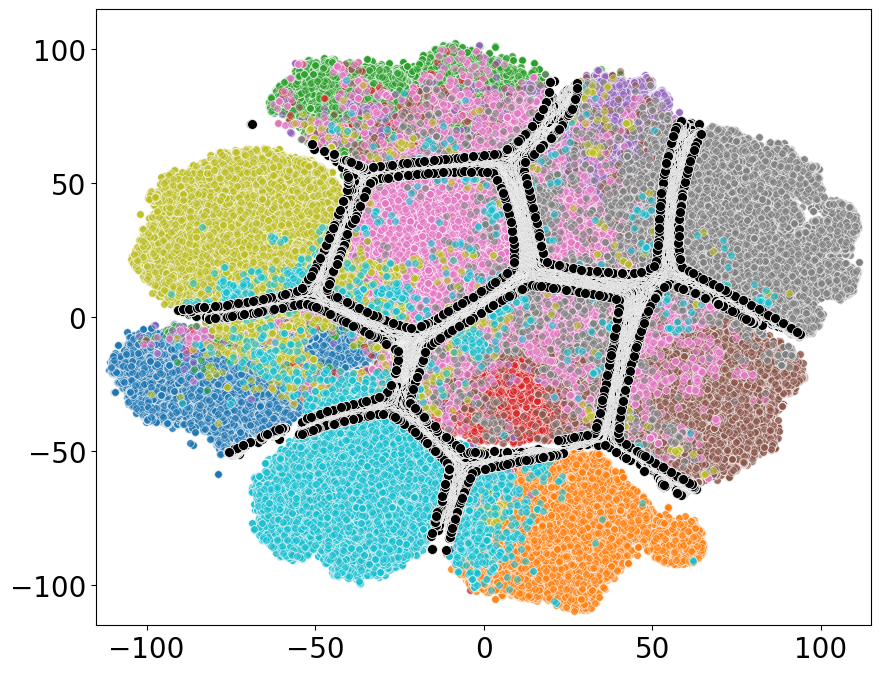}
        \label{fig:gmm}
    }\hfill
    \subfloat[LCS-KM]{
        \includegraphics[width=.31\linewidth]{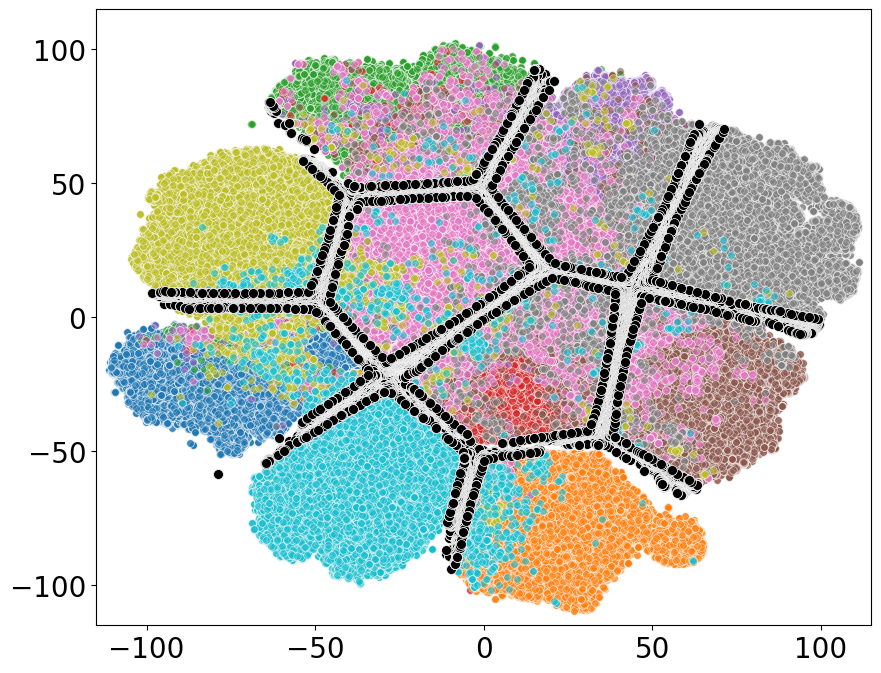}
        \label{fig:kmeans}
    }
    \vspace{-0.05in}
    \caption{Visual comparison of selection techniques on TinyImages dataset in the latent space. Each subplot represents a different method: (a) PCS identifies points with the lowest classification confidence, highlighting areas where the model is most uncertain within the ten-class latent representation, (b) LCS-GMM illustrates probability contours from Gaussian Mixture Models, with selected points emphasizing regions of overlapping probabilities among the $10$ class clusters, and (c) LCS-KM highlights points selected near decision boundaries across $10$ classes based on k-means clustering in the latent space.}
    \vspace{-0.1in}
    \label{fig:clustering_techniques}
\end{figure*}

\section{Further Analyses}
\label{sec:additional analysis}

This section presents further analyses, including evaluations on a real-world medical dataset (Section \ref{app:medical}), visualizations of selected unlabeled data (Section \ref{sec:insight of our method}), and ablation studies on key hyperparameters (Section \ref{sec:hyperparameter tuning}).
Due to the page limit, additional experiments under different training schemes and perturbations, as well as the effect of the intermediate model in low-label regimes, are provided in Appendix~\ref{app:more_analysis}.

\subsection{Application to Real-World Medical Data}
\label{app:medical}

We extend our study to a real-world healthcare application. COVIDGR~\cite{tabik2020covidgr} is used as the labeled dataset, and CoronaHack~\cite{govi2020coronahack} serves as the source of the unlabeled data. 
Specifically, COVIDGR contains $852$ chest X-ray images, evenly distributed across two classes: 
$426$ COVID-19 positive (RT-PCR confirmed) and $426$ negative cases. It is then split into $80\%$ training and $20\%$ test sets. CoronaHack comprises $5910$ images with an imbalanced class distribution ($73\%$ positive, $27\%$ negative), 
split into an $89\%$ training set and an $11\%$ test set. 
We consider $\ell_\infty$ perturbations with $\epsilon=0.1$ and train models on a ResNet-18 architecture using SSAT by applying our LCS-KM strategy to select unlabeled CoronaHack data with varying ratios (see Appendix \ref{append:experimental details_medicalapplication} for more details).

Figure \ref{fig:four_graphs} demonstrates the results, which again confirm the effectiveness of our selection scheme. For instance, using LCS-KM to select $10\%$ of the unlabeled data achieves a robust accuracy of $56\%$, compared to $53\%$ when the same proportion is chosen randomly (Figures \ref{fig:10perrandom} and \ref{fig:10per}). In comparison, using a smaller unlabeled dataset leads to faster convergence, underscoring the benefits of early stopping. Specifically, the best performance of $56\%$ is reached at epoch $36$ when using $10\%$ of the data selected with LCS-KM (Figure~\ref{fig:10per}). 
Figure~\ref{fig:20per} shows that incorporating $20\%$ of the data with LCS-KM improved the accuracy to $57\%$ at epoch $47$. In contrast, training on the entire unlabeled dataset requires $60$ epochs to achieve a peak robust accuracy of $58\%$ (Figure~\ref{fig:100per}).
These findings confirm that using a smaller, but strategically selected, unlabeled data subset can achieve robustness comparable to that of the entire dataset. These results highlight the practical relevance of our data reduction scheme, particularly for applications where labeled data is limited yet robust model performance is essential.

\begin{figure*}[t] 
\vspace{-0.1in}
    \centering
    \subfloat[CIFAR-10 ($\alpha=20\%$)]{
        \includegraphics[height=0.15\linewidth, width=0.235\linewidth]{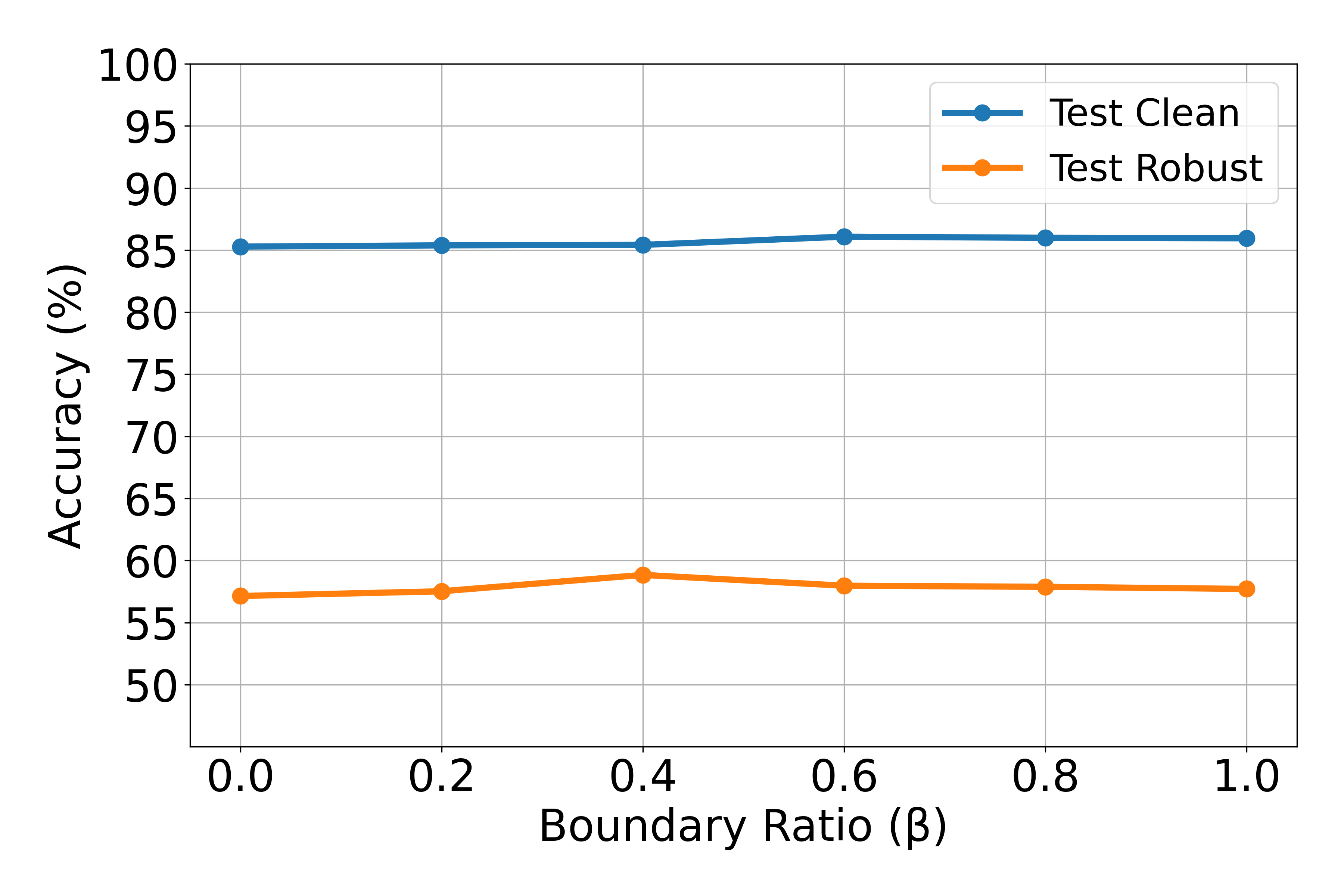}
        \label{fig:betacifar}
    }
    \subfloat[SVHN ($\alpha=10\%$)]{
        \includegraphics[height=0.15\linewidth, width=0.235\linewidth]{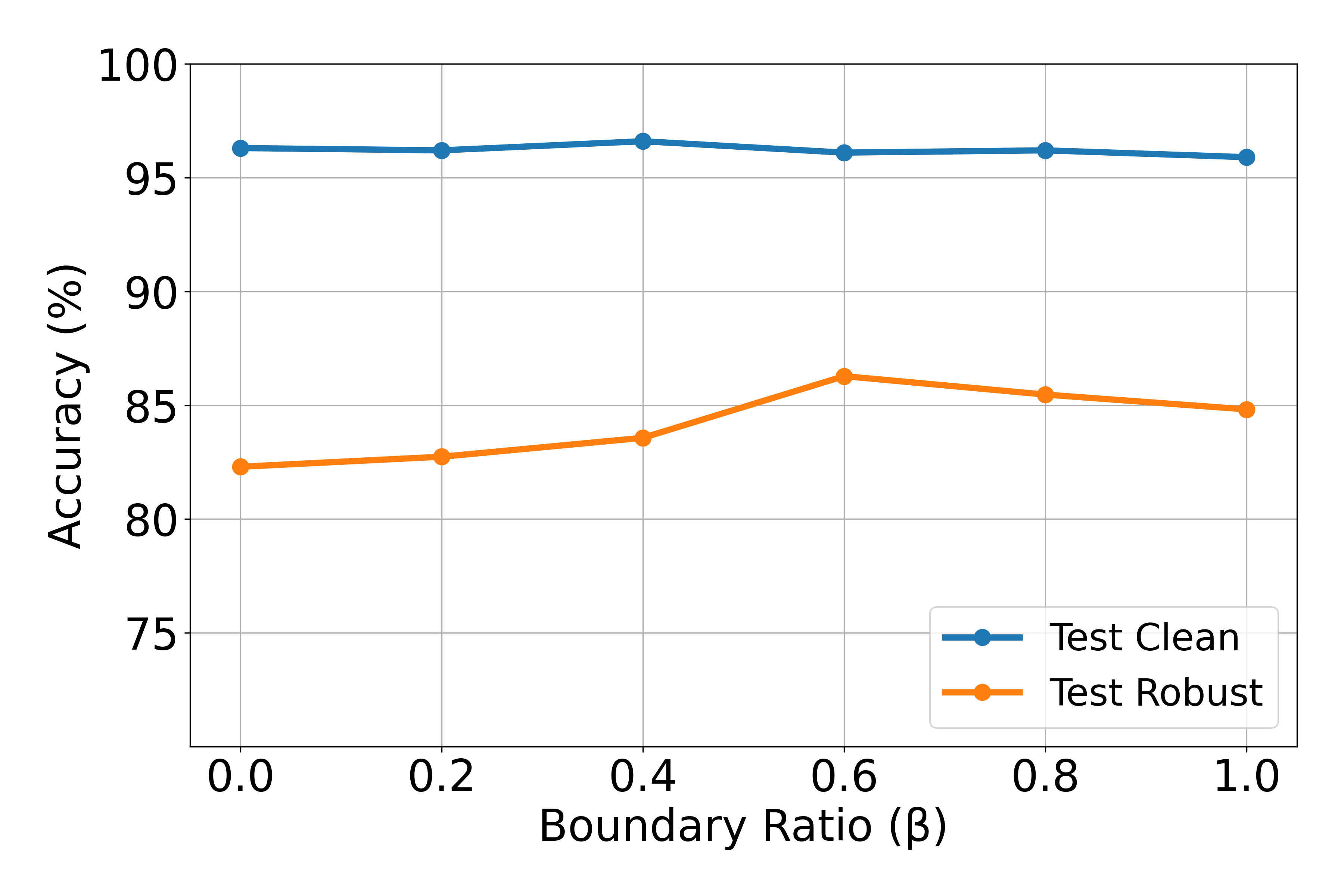}
        \label{fig:betasvhn}
    }
    \subfloat[Fine-Tuning Epochs $S$]{
        \includegraphics[height=0.15\linewidth, width=0.235\linewidth]{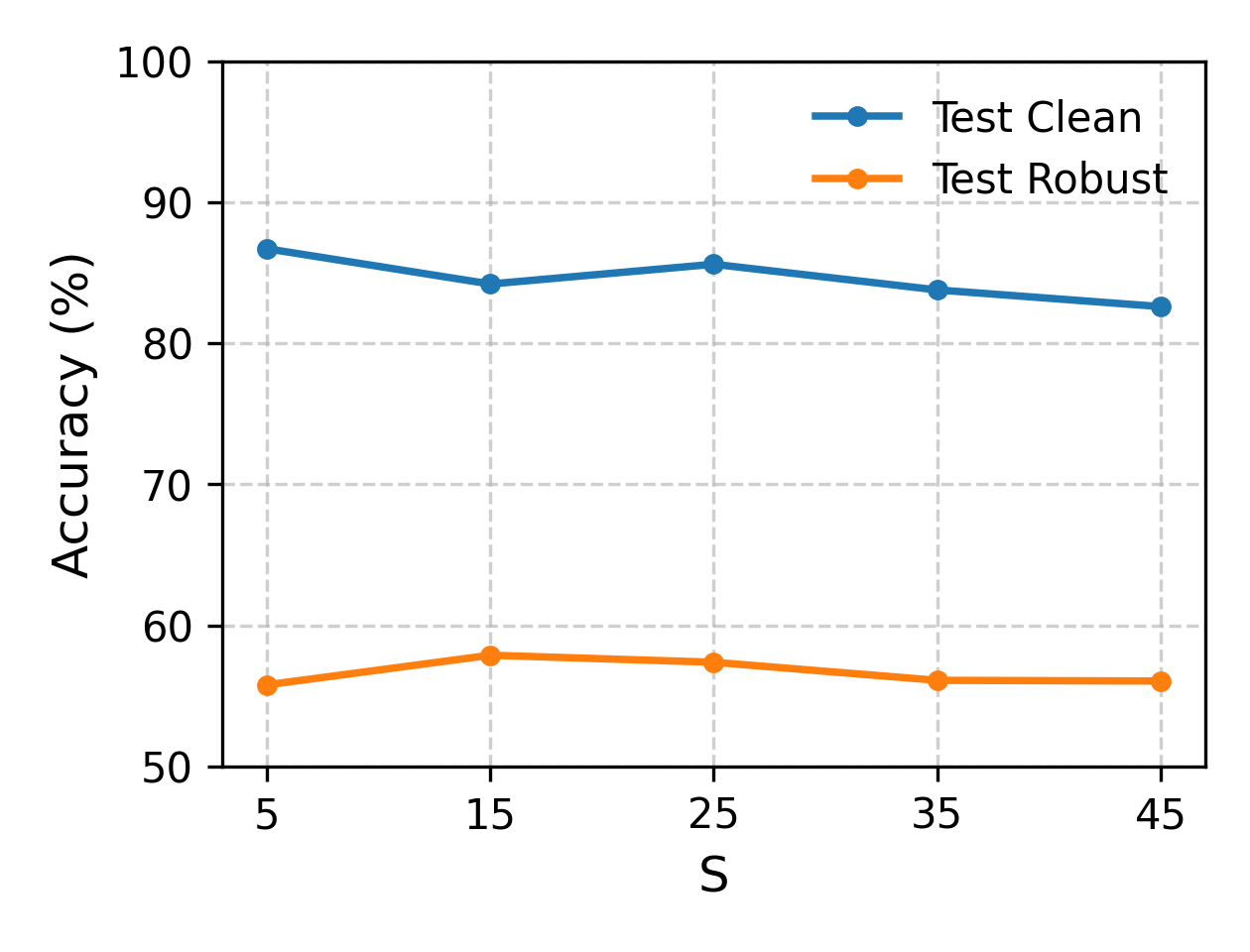}
        \label{fig:finetuneepochvsaccuracy}
    }
    \subfloat[Regularization Strength $\lambda$]{
        \includegraphics[height=0.15\linewidth, width=0.235\linewidth]{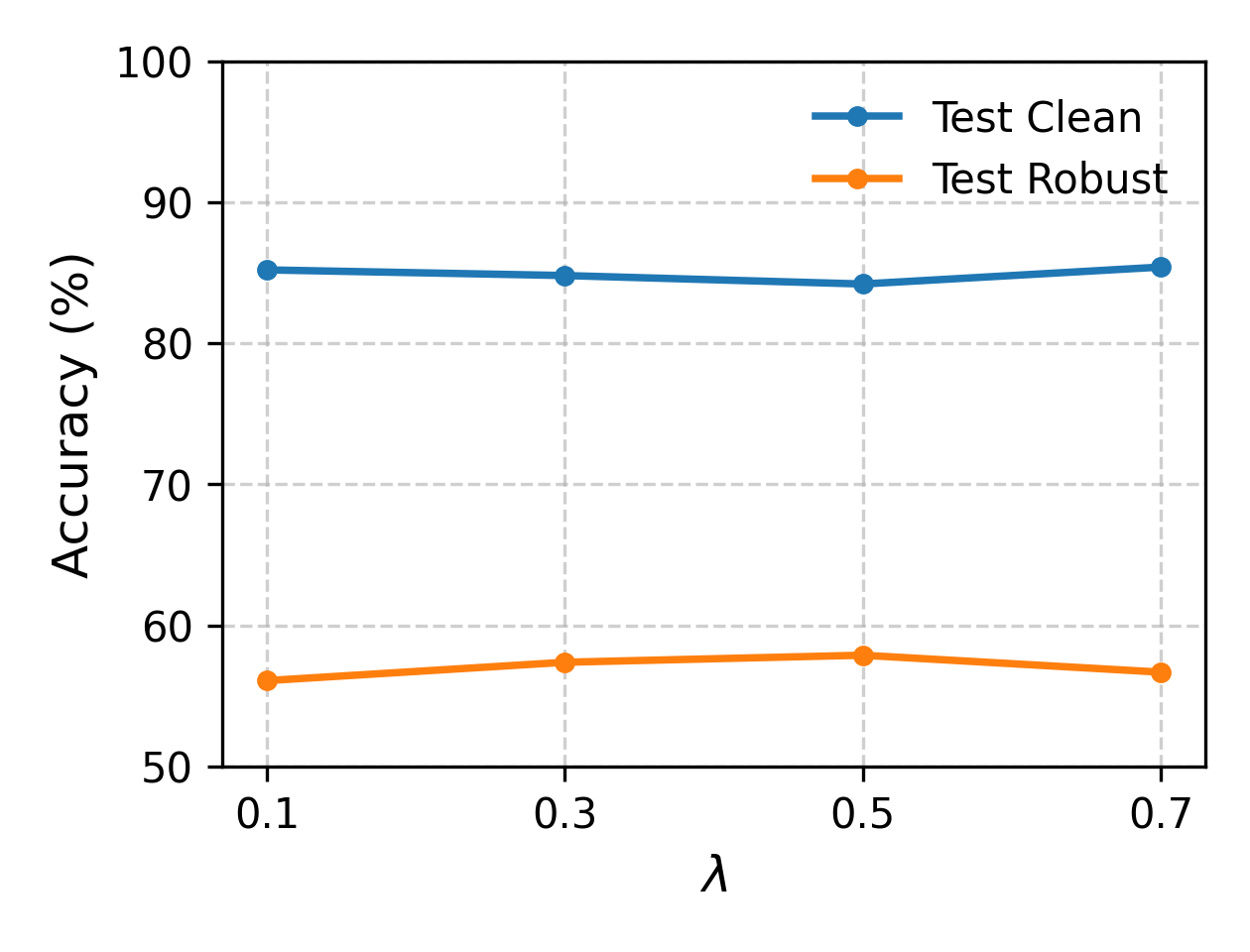}
        \label{fig:lambdavsaccuracy}
    }
    \vspace{-0.1in}
    \caption{Figures (a) and (b): Illustration of SSAT performance using LCS-KM by varying boundary ratio parameters $\beta$ with respect to DDPM-generated CIFAR-10 data with $\alpha = 20\%$ and $531$K extra SVHN data with $\alpha = 10\%$.  Figures (c) and (d): Clean and robust accuracy curves of SSAT with LCG-KM on CIFAR-10 with $\alpha = 10\%$ by varying: (a) the number of fine-tuning epochs $S$, and (b) the regularization strength parameter $\lambda$.}
    \vspace{-0.1in}
\end{figure*}

\subsection{Comparative Visualization}
\label{sec:insight of our method}

To facilitate a deeper understanding, we perform a comparative visualization of the selected unlabeled data on CIFAR-10. 
We first extract the latent representations from the penultimate layer of a pre-trained WRN-28-10 intermediate model. 
For visualization, we employ a two-step dimension reduction that balances the retention of global structure and the visualization of local relationships. 
In particular, we first perform \emph{Principal Component Analysis} (PCA)~\cite{gewers2021principal}. By reducing the dimensionality of the latent features to $50$, PCA retains the majority of the variance within the data. 
Then, we apply \emph{t-Distributed Stochastic Neighbor Embedding} (t-SNE)~\cite{xia2021revisiting} to project the $50$-dimensional PCA features onto a $2$-dimensional space. Unlike PCA, which emphasizes preserving variance, t-SNE mainly aims to maintain the local relationships among data points. 
Finally, we visualize the set of unlabeled data selected by our methods under $\alpha=10\%$.
Figure \ref{fig:clustering_techniques} shows the visualization results, where each color represents the label class predicted by the intermediate model for the corresponding unlabeled point. The highlighted black dots indicate the data points selected by the respective selection strategy. 

Specifically, PCS adopts a purely confidence-based selection strategy, choosing the points with the globally lowest confidence without considering their distribution within or across classes. 
Although this approach is appealing for its simplicity, it can inadvertently prioritize noisy outliers.
As the number of classes grows and data complexity increases, this scattershot selection tends to yield a more disorganized set of points (Figure \ref{fig:PCS_selection}). 
In comparison, LCS-GMM and LCS-KM yield more structured sampling patterns.
While LCS-GMM can roughly capture regions where classes overlap, it assumes that each cluster follows approximately Gaussian distributions, an assumption that may fail in real-world settings. As a result, LCS-GMM’s contours become misaligned with the true underlying structure, leading to less accurate localization of essential boundary points (Figure \ref{fig:gmm}). A closer look at the selected samples also reveals that the corresponding region is broader and less concentrated on the decision boundary, making it harder to capture truly critical samples for SSAT. 
Our best-performing LCS-KM method emphasizes the importance of leveraging the latent-space clustering to identify boundary-adjacent data points in a more accurate and balanced manner. By partitioning the latent embeddings into multiple k-means clusters, LCS-KM is able to locate more points with low classification confidence within each cluster (Figure \ref{fig:kmeans}). 
Compared with LCS-GMM, a notable pattern is that the unlabeled data picked by LCS-KM are aligned more linearly.
This coincides with the piecewise linear characteristic of neural networks' decision boundary, especially for the last layer.
We hypothesize that this local linear property may explain why our latent clustering-based approach with k-means consistently achieves the best performance compared to the other two alternatives.

\subsection{Hyperparameter Sensitivity}
\label{sec:hyperparameter tuning}

We conduct ablations to study the sensitivity of the hyperparameters involved in our methods
The detailed setup and the ablations on the unlabeled data ratio $\alpha$ and the per-batch ratio of extra to original data $\gamma$ are deferred to Appendix \ref{append:ratio}.

\shortsection{Ratio of Boundary Data $\bm\beta$}
Concentrating exclusively on data points near the decision boundary can result in an excessive number of points close to the boundary and too few farther away, leading to overfitting. To mitigate this, we choose a combination of points, with some near the boundary and others further away, where the ratio is determined by \(\beta\). 
Figure \ref{fig:betacifar} illustrates how robust accuracy varies with \(\beta\) on CIFAR-10 using $20\%$ generated data from DDPM. In this scenario, the optimal robust accuracy is achieved when \(\beta\) is $0.4$, meaning $40\%$ of the selected points are near the boundary, while $60\%$ are farther away. 
Figure \ref{fig:betasvhn} illustrates how robust accuracy varies with \(\beta\) on SVHN using $10\%$ extra data from the same dataset. Due to a smaller amount of external data, selecting points near the boundary results in a lower risk of overfitting. As a result, the highest robust accuracy is observed when 
\(\beta\) is $0.6$ with  $60\%$ of the points near the boundary and $40\%$ farther away.
Therefore, the ideal value of \(\beta\) depends on the amount of external data added and the ratio of external to original data. When more external data is added, using a smaller \(\beta\) typically yields better robust accuracy. 
Conversely, choosing a larger \(\beta\) is favorable when there is less extra unlabeled data.

\shortsection{Guided DDPM Fine-Tuning ($S$ \& $\lambda$)}
We conduct ablations on our best-performing LCG-KM approach by varying the number of fine-tuning epochs $S$ and the regularization strength parameter $\lambda$ on CIFAR-10 under $\alpha = 10\%$. 
Figure \ref{fig:finetuneepochvsaccuracy} illustrates our ablation results on $S$.
We vary the number of fine-tuning epochs $S\in\{5, 15, 25, 35, 45\}$ with $\lambda = 0.5$, and report the final clean and robust accuracies after the whole SSAT process. 
We observe that choosing a small number of epochs for fine-tuning DDPM (e.g., $10$--$20$ epochs) is usually sufficient to achieve the best robustness performance, likely because the starting DDPM is already pre-trained on CIFAR-10. 
Training for too many epochs can lead to a degradation in image quality 
and loss of diversity (see Figure \ref{fig:generative_visual} in the appendix for visualizations), thereby adversely influencing the final robust accuracies.
Figure \ref{fig:lambdavsaccuracy} shows the results for the regularization strength parameter $\lambda\in\{0.1, 0.3, 0.5, 0.7\}$ with fixed epoch number $S = 15$.
Our findings indicate that selecting $\lambda$ within the range of $[0.4, 0.6]$ yields the best performance, striking a balance between the quality of the generated images and their proximity to the model's decision boundary.

\section{Conclusion and Future Work}
\label{sec:conclusion}

In this work, we developed various data reduction schemes to improve the efficiency of SSAT.
We illustrated that strategically selecting or generating more boundary-adjacent data points can significantly improve the data efficiency of SSAT while preserving its robustness advantages.
In particular, our latent-based k-means clustering approach performs the best, reducing both computational costs and memory requirements.

\shortsection{Limitation \& Future Direction}
While LCS-KM consistently achieves the best performance, its success requires careful hyperparameter tuning; future work can leverage AutoML algorithms to automate this process.
In addition, theoretical analysis of how to solve the optimization problems (Equations \ref{eq:problem formulation} and \ref{eq:problem formulation generation}) in a more principled way would be helpful. Finally, extending our guided DDPM fine-tuning with training-free sampling-based guidance~\cite{dhariwal2021diffusion} or advanced generative models~\cite{song2020denoising,geng2025mean} is a promising direction. Further discussions are provided in Appendix \ref{append:further discussions}.
Despite these challenges, we believe our work can serve as an important initial step towards more efficient and scalable development of robust models.



\bibliographystyle{IEEEtran}
\bibliography{ref}

\appendices

\section{Algorithm Pseudocode}
\label{append:alg}

Algorithms \ref{alg:dataselection_confidence}-\ref{alg:diffusion_finetune} present the pseudocode of our data reduction schemes, corresponding to prediction confidence-based selection (PCS), latent clustering-based selection (LCS), and guided DDPM-based fine-tuning, as introduced in Section \ref{sec:method}.

\section{Detailed Experimental Setup}
\label{append:experimental details}

\shortsection{SVHN}
SVHN is naturally divided into a core training set comprising approximately $73$K images and an extra training set of around $531$K  images. Initially, the model is trained on the $73$K labeled images.   
To evaluate the generalizability of our selection schemes, we apply them to a synthetic dataset generated using a DDPM model \cite{gowal2021improving}.

\shortersection{Configuration}
For model training, we adopt a WideResNet 16-8 (WRN-16-8) architecture \cite{zagoruyko2016wide} in all our SVHN experiments. We generate adversarial examples using PGD attacks exactly as implemented by Zhang et al. \cite{zhang2019theoretically}, with step size $0.007$, $10$-step PGD attack iterations, and $\ell_\infty$ perturbation magnitude $\epsilon = 0.015$.
Hyperparameters are set the same as in Carmon et al. \cite{carmon2019unlabeled} except for the number of epochs. For SVHN, we use a training batch size of $128$.
All of our experiments that utilize all of the extra 531K data are run for $200$ epochs, and the experiments with $1$M generated data are run for $400$ epochs. To prevent overfitting and reduce computational complexity, we run our experiments with selected data for $75$ epochs with early stopping.
Following TRADES~\cite{zhang2019theoretically}, we employ an SGD optimizer with a weight decay factor of $5\cdot 10^4$ and an initial learning rate of $0.1$, along with a stepwise learning rate scheduler with a decaying factor of $10$ at epochs $75$ and $90$.

\begin{algorithm}[t]
\caption{Prediction Confidence-Based Selection (PCS)}
\label{alg:dataselection_confidence}
\setstretch{1.1}
\small
\begin{algorithmic}[1]
\STATE {\bfseries Input:} labeled $\mathcal{S}_l$; unlabeled $\mathcal{S}_u$; selection ratio $\alpha$; perturbation size $\epsilon$; boundary ratio $\beta$; per-batch ratio $\gamma$
\STATE $\hat{\theta}\leftarrow$ train a standard classification model on $\mathcal{S}_l$
\STATE $\hat{y} \leftarrow$ predict pseudo label $f_{\hat{\theta}}(\bm{x})$ for $\bm{x}\in \mathcal{S}_u$
\STATE Initialize $
\mathcal{A}_u$ as empty set
\STATE $\mathrm{Conf}(\bm{x}) \leftarrow$ get confidence of $f_{\hat\theta}(\bm{x})$ for $\bm{x} \in \mathcal{S}_u$ 
\STATE Sort $\mathcal{S}_u$ by ascending order of $\mathrm{Conf}(\bm{x})$ 
\STATE $\mathcal{A}_u \leftarrow$ add top $\beta \cdot \alpha|\mathcal{S}_u|$ points with lowest $\mathrm{Conf}(\bm{x})$
\STATE $\mathcal{A}_u \leftarrow$ add $(1-\beta) \cdot \alpha|\mathcal{S}_u|$ points randomly from $\mathcal{S}_u \setminus \mathcal{A}_u$
\STATE ${\theta}_\mathrm{final}\leftarrow$ $\mathrm{SSAT}(\mathcal{S}_l, \mathcal{A}_u, \gamma)$ 
\STATE {\bfseries Output:} selected subset $\mathcal{A}_u$, final model ${\theta}_\mathrm{final}$
\end{algorithmic}
\end{algorithm}

\begin{algorithm}[t]
\caption{Latent Clustering-Based Selection (LCS)}
\label{alg:dataselection_clustering}
\setstretch{1.1}
\small
\begin{algorithmic}[1]
\STATE {\bfseries Input:} labeled $\mathcal{S}_l$; unlabeled $\mathcal{S}_u$; selection ratio $\alpha$; cluster number $k$; perturbation size $\epsilon$; boundary ratio $\beta$; per-batch ratio $\gamma$
\STATE $\hat{\theta}\leftarrow$ train a standard classification model on $\mathcal{S}_l$
\STATE $\hat{y} \leftarrow$ predict pseudo label $f_{\hat{\theta}}(\bm{x})$ for $\bm{x}\in \mathcal{S}_u$
\STATE Initialize $
\mathcal{A}_u$ as empty set
\STATE $\bm{z}\leftarrow$ get latent embeddings $h_{\hat{\theta}}(\bm{x})$ for $\bm{x} \in \mathcal{S}_u$

\IF{using LCS-KM}
    \STATE $\{\mathcal{C}_1, \ldots, \mathcal{C}_k\} \leftarrow$ k-means clustering on $\{\bm{z}\}_{\bm{x}\in\mathcal{S}_u}$
    \STATE $\Delta d \leftarrow$ compute $\Delta d = |d_1 - d_2|$ for $\bm{x}\in \mathcal{S}_u$, where $d_1$ and $d_2$ are Euclidean distances to the nearest two centroids
    \STATE $\mathcal{A}_u \leftarrow$ add top $\beta \cdot \alpha|\mathcal{S}_u|$ points with smallest $\Delta d$ 
\ELSIF{using LCS-GMM}
    \STATE Fit a GMM with $k$ components to $\{\bm{z}\}_{\bm{x}\in\mathcal{S}_u}$
    \STATE Compute GMM posterior probability $\bm{p}$ for each $\bm{z}$
    \STATE $\Delta p \leftarrow$ compute $\Delta p = |p_1 - p_2|$ for each $\bm{z}$, where $p_1$ and $p_2$ are the top two highest probabilities
    \STATE $\mathcal{A}_u \leftarrow$ add top $\beta \cdot \alpha|\mathcal{S}_u|$ points with smallest $\Delta \gamma$ 
\ENDIF
\STATE $\mathcal{A}_u \leftarrow$ add $(1-\beta) \cdot \alpha|\mathcal{S}_u|$ points randomly from $\mathcal{S}_u \setminus \mathcal{A}_u$
\STATE ${\theta}_\mathrm{final}\leftarrow$ $\mathrm{SSAT}(\mathcal{S}_l, \mathcal{A}_u, \gamma)$ 
\STATE {\bfseries Output:} selected dataset $\mathcal{A}_u$, final model ${\theta}_\mathrm{final}$
\end{algorithmic}
\end{algorithm}

\begin{algorithm}[t]
\caption{Guided DDPM Fine-Tuning}
\label{alg:diffusion_finetune}
\setstretch{1.15}
\small
\begin{algorithmic}[1]
\STATE {\bfseries Input:} labeled $\mathcal{S}_l$; pre-trained DDPM $\theta_{\mathrm{pre}}$; fine-tuning epochs $S$; trade-off parameter $\lambda$; other parameters $\epsilon$, $\alpha |\mathcal{S}_u|$, $\beta$, $\gamma$, $k$, $\eta$
\STATE $\hat{\theta}\leftarrow$ train a standard classification model on $\mathcal{S}_l$
\STATE Initialize $\mathcal{G}_u$ as empty set 
\STATE $\mathcal{G}_u \leftarrow$ add $(1-\beta) \cdot \alpha |\mathcal{S}_u|$ points generated by pre-trained $\theta_{\mathrm{pre}}$
\STATE Initialize $\theta$ as $\theta_{\mathrm{pre}}$
\FOR{$s = 1, 2, \ldots, S$}
    \STATE  Sample batches of size $b$: $\bm{x}_0\sim\mathcal{S}_l$, $t \sim \mathcal{U}(T)$, $\bm{\epsilon} \sim \mathcal{N}(\bm{0}, \mathbf{I})$
    \STATE $\bm{x}_t \leftarrow \sqrt{\bar{\alpha}_{t_i}} \bm{x}_0 + \sqrt{1 - \bar{\alpha}_{t}} \bm{\epsilon}$ 
    \STATE $ L_{\mathrm{DDPM}} \leftarrow \frac{1}{b}\sum_{i\in[b]}\| \bm{\epsilon} - \bm{\epsilon}_{\theta}(\bm{x}_{t}, t) \|_2^2$
    \IF{using PCG}
        \STATE $L_{\mathrm{reg}} \leftarrow \frac{1}{b}\sum_{i\in[b]}\| \ell_{\mathrm{PC}}(g_{\theta} (\bm{x}_t, t)) \|_2^2$
    \ELSIF{using LCG-KM}
        \STATE $L_{\mathrm{reg}} \leftarrow \frac{1}{b}\sum_{i\in[b]}\| \ell_{\mathrm{KM}}(g_{\theta} (\bm{x}_t, t)) \|_2^2$
    \ELSIF{using LCG-GMM}
        \STATE $L_{\mathrm{reg}} \leftarrow \frac{1}{b}\sum_{i\in[b]}\| \ell_{\mathrm{GMM}}(g_{\theta} (\bm{x}_t, t)) \|_2^2$
    \ENDIF
    \STATE $\theta \leftarrow \theta - \eta \nabla_{\theta} \big( L_{\mathrm{DDPM}} +  \lambda \cdot L_{\mathrm{reg}} \big)$
\ENDFOR
\STATE $\mathcal{G}_u \leftarrow$ add $\beta \cdot \alpha |\mathcal{S}_u|$ points generated by fine-tuned $\theta$
\STATE ${\theta}_\mathrm{final}\leftarrow$ $\mathrm{SSAT}(\mathcal{S}_l, \mathcal{G}_u, \gamma)$
\STATE {\bfseries Output:} generated dataset $\mathcal{G}_u$, final model $\theta_{\mathrm{final}}$
\end{algorithmic}
\end{algorithm}

\shortersection{Evaluation} 
For the attack evaluation to calculate the robust accuracy using PGD attack, we keep the parameters similar to those of Carmon et al. \cite{carmon2019unlabeled} for better comparison. We use a step size $0.005$, and a number of attack steps of $K = 100$. We evaluate models at $\epsilon = 0.015$, which is the same value used during training. We also use Auto Attack with the same $\epsilon$ value as in the PGD attack.

\shortsection{CIFAR-10}
The CIFAR-10 dataset has 50K labeled images. We use the $80$ Million Tiny Images (80M-TI) dataset, of which CIFAR-10 is a manually labeled subset, to obtain extra data. However, most images in $80$M-TI do not correspond to CIFAR-10 image categories. Carmon et al. \cite{carmon2019unlabeled} used an 11-way classifier to distinguish CIFAR-10 classes and an 11th ``non-CIFAR-10'' class using a WideResNet 28-10 model. Each class selected an additional $50$K images from 80M-TI using the model’s predicted scores to create a $500$K pseudo-labeled dataset, which we used in our experiments. We train the intermediate model using $50$K labeled data and utilize this model to select data from either the $500$K pseudo-labeled data or a synthetic dataset. We perform the same experiments as in SVHN, except that we do not have the ground-truth labels of the additional unlabeled dataset. We also conducted further experiments using $1$M synthetic data generated by the DDPM model \cite{gowal2021improving}. For our fine-tuning experiments, we start with a publicly available pretrained DDPM model \cite{ho2020ddpm_cifar10_32} and further fine-tune it for a few epochs to generate the desired images, which are subsequently used for our final training.

\shortersection{Configuration} For our main CIFAR-10 experiments, we use a WideResNet 28-10 (WRN-28-10) \cite{zagoruyko2016wide} architecture to train models.
 Similar to SVHN experiments, we set step size $0.007$, PGD attack iterations as $10$, and $\ell_\infty$ perturbation magnitude $\epsilon = 0.031$.
 For our adversarial training with $\ell_2$ perturbation, we use $\epsilon = 128/255$. In section \ref{sec:exploration of generalizability}, when $\epsilon = 4/255$, we use $0.0035$ as the PGD step size. For $\epsilon = 16/255$, we use $0.015$ as the PGD step size.
 Hyperparameters used are the same as in Carmon et al. \cite{carmon2019unlabeled} except for the number of epochs. Here, we use a training batch size of $256$.
 The experiments using all the $500$K pseudo-labeled data are run for $200$ epochs, and using 1M generated data are run for $400$ epochs. 
 For our experiments with limited data, we get the best results at $100$ epochs, where we early stop the training process. 
 We employ an SGD optimizer with a weight decay factor of $5\cdot 10^4$ and an initial learning rate of $0.1$, along with a stepwise scheduler with a decaying factor of $10$ at epochs $75$ and $90$, respectively.

\shortersection{Evaluation} 
The PGD attack evaluation is conducted similarly to that of Carmon et al. \cite{carmon2019unlabeled} for fair comparisons. We use a step size $\alpha = 0.01$, and a number of attack steps $K = 40$. We consider $\ell_\infty$ perturbations with $\epsilon = 0.031$, the same as the perturbation magnitude we used during training. For the models trained with $\ell_2$ perturbation, we evaluate with $\ell_2$ perturbation magnitude $\epsilon = 128/255$. We also use Auto Attack with the same $\epsilon$ value as in the PGD attack.

\shortsection{Medical Application}
\label{append:experimental details_medicalapplication}
For experiments on the medical dataset outlined in Section \ref{app:medical}, we train models using a ResNet-18  architecture \cite{he2016deep}, and consider $\ell_\infty$ perturbations with $\epsilon = 0.1$. 
We set the attack step size to $0.02$, and the number of PGD steps to $10$.
The final model is trained using SSAT with the Adam optimizer \cite{kingma2014adam} and a learning rate of $0.001$.
We evaluate model robustness with PGD attacks with a step size of $0.02$ and $20$ attack steps.

\begin{figure}[!t]
    \centering
    \includegraphics[width=1.0\linewidth]{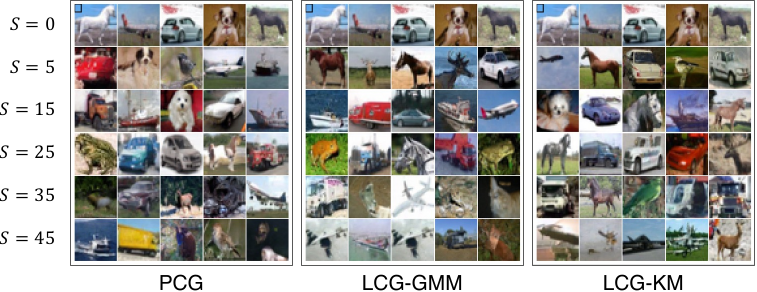} 
    \vspace{-0.25in}
    \caption{Visualization of generated images with various fine-tuning epochs $S$.}
    \vspace{-0.05in}
    \label{fig:generative_visual}
\end{figure}

\section{Additional Experiments}
\label{append:additional experiments}

\begin{figure*}[t] 
\vspace{-0.1in}
    \centering
    \subfloat[CIFAR-10]{
        \includegraphics[width=0.235\linewidth]{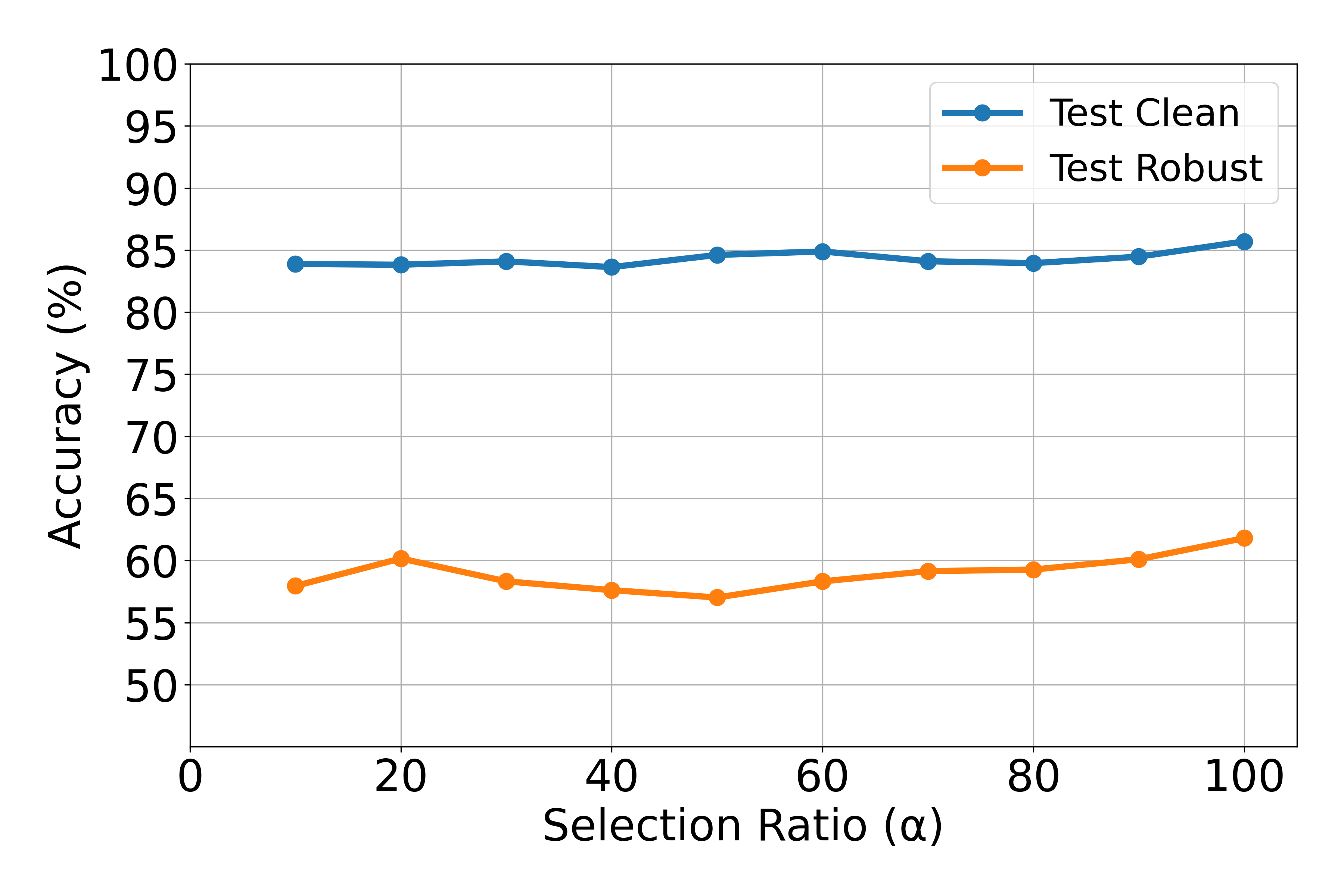}
        \label{fig:cofarper}
    }
    \subfloat[SVHN]{
        \includegraphics[width=0.235\linewidth]{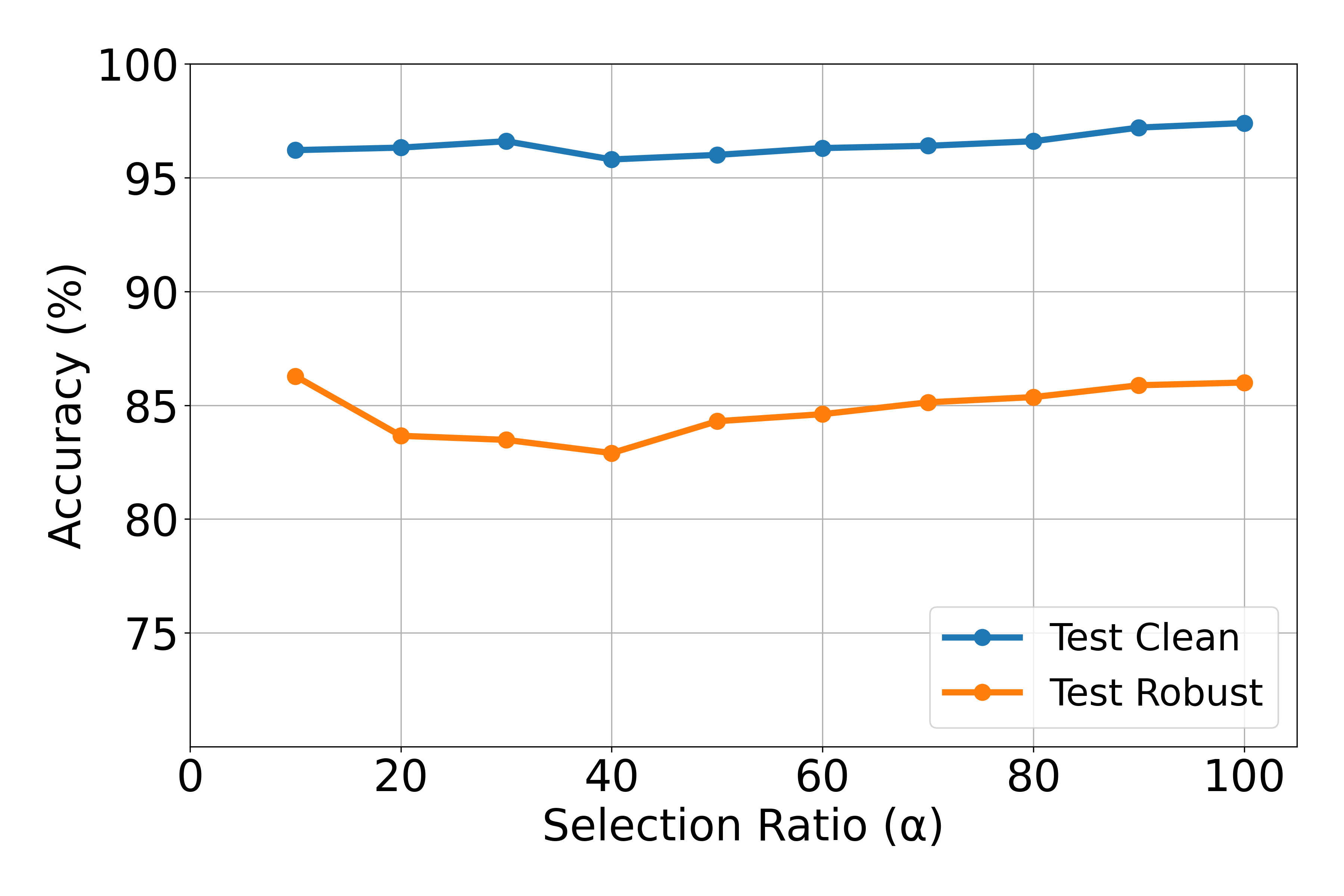}
        \label{fig:svhnper}
    }
    \subfloat[CIFAR-10]{
    \includegraphics[width=0.235\linewidth]{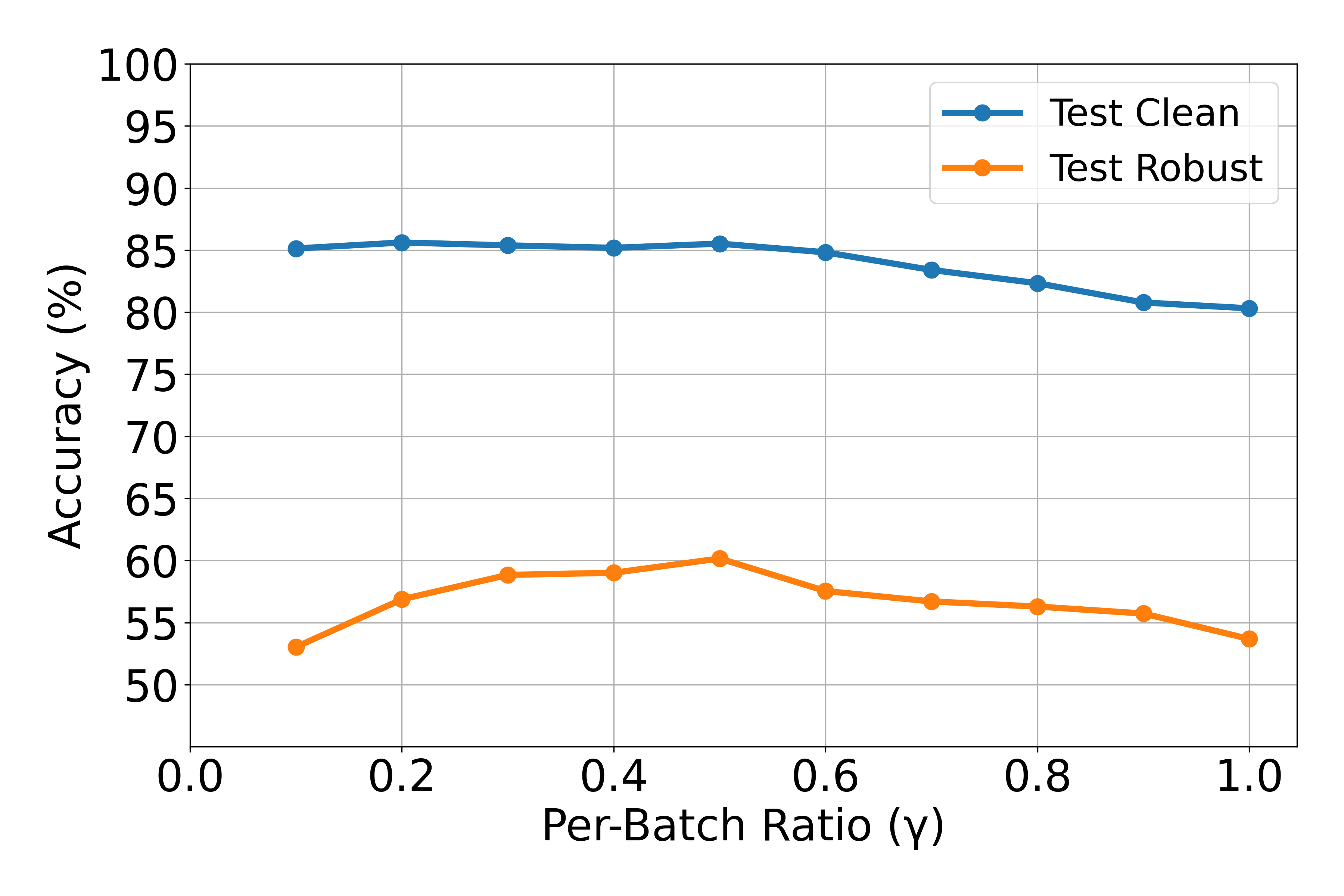}
    \label{fig:ratiocifar}
    }
    \subfloat[SVHN]{
        \includegraphics[width=0.235\linewidth]{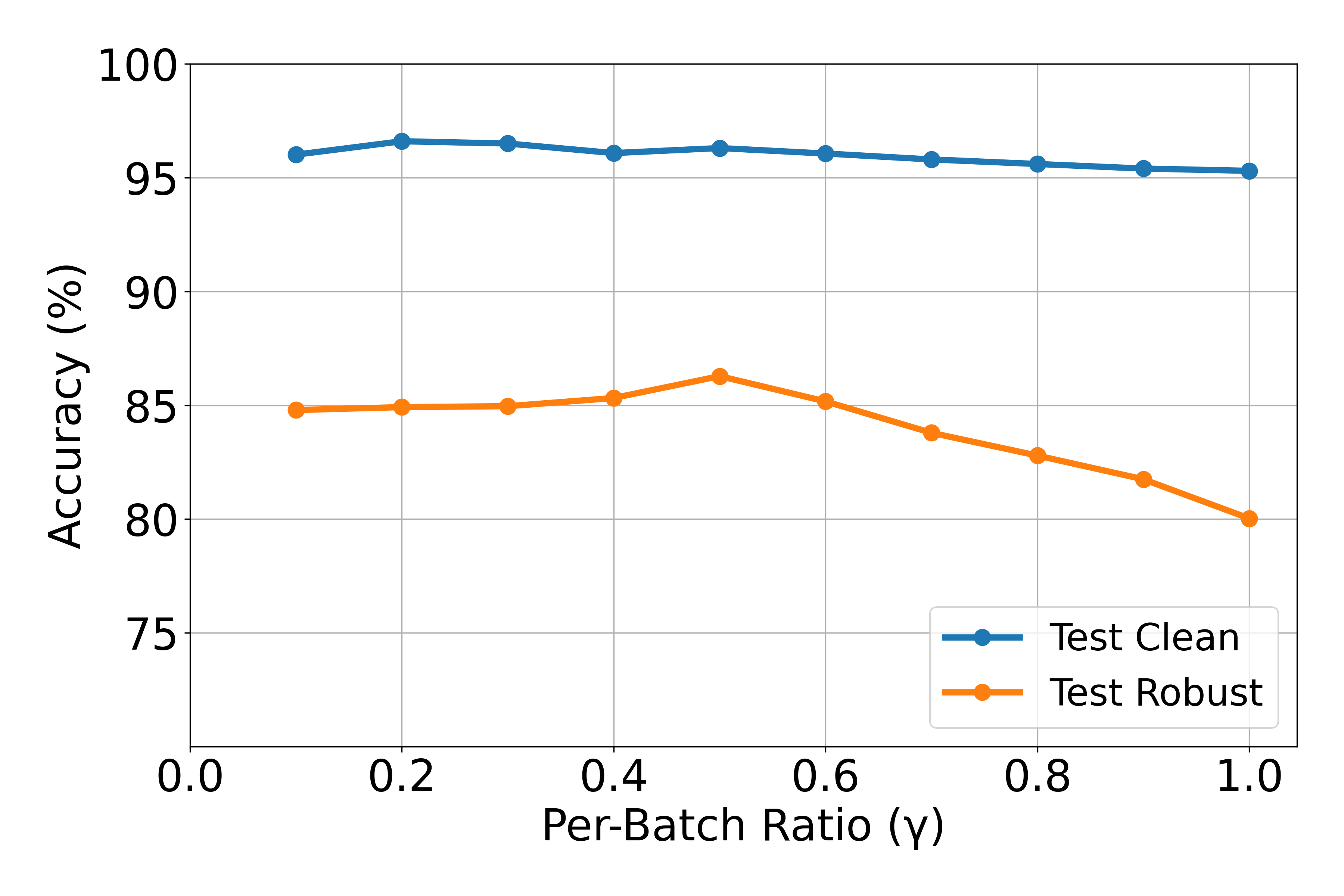}
        \label{fig:ratiosvhn}
    }
    \vspace{-0.1in}
    \caption{Figures (a) and (b): Illustration of SSAT performance using LCS-KM by varying unlabeled data ratios $\alpha$ for CIFAR-10 with DDPM generation ($\beta = 0.4$) and SVHN with external data ($\beta = 0.6$). Figures (c) and (d):
    Clean and robust accuracy curves of SSAT with LCS-KM by varying per-batch ratio $\gamma$ on CIFAR-10 with $1$M generated images using DDPM ($\alpha=20\%$, $\beta=0.4$) and on SVHN with external $531$K images ($\alpha=10\%$, $\beta=0.6$).
    }
\end{figure*}

\subsection{Other Ablations on Hyperparameters}
\label{append:ratio}

In Section \ref{sec:hyperparameter tuning}, we present a detailed analysis of how accuracies are influenced by different hyperparameter values. We vary one hyperparameter at a time while keeping the others constant. Figure \ref{fig:betacifar} illustrates how robust accuracy on the CIFAR-10 dataset changes with \(\beta\) values when $20\%$ additional data is incorporated from the $1$M data samples generated by the DDPM model, using a per-batch ratio of $0.3$. Figure \ref{fig:betasvhn} shows the variation in robust accuracy for the SVHN dataset as \(\beta\) changes, with $10\%$ drawn from the SVHN extra dataset and a per-batch ratio of $0.5$. For our LCG ablations on $S$ and $\lambda$ (Figures \ref{fig:finetuneepochvsaccuracy} and \ref{fig:lambdavsaccuracy}), we set $\alpha=20\%$, $\beta=0.4$ and $\gamma=0.3$ for CIFAR-10, and choose $\alpha=10\%$, $\beta=0.6$ and $\gamma=0.5$ for SVHN, matching the same set of optimal hyperparameters as identified for our selection-based strategies. Figure \ref{fig:generative_visual} visualizes the images generated by our guided DDPM fine-tuning techniques, including PCG, LCG-GMM, and LCG-KM, on CIFAR-10 with different numbers of fine-tuning epochs $S\in\{0,5,15,25,35,45\}$. As $S$ is increased, the generated CIFAR-10 images become more difficult to recognize, reflecting a higher degree of uncertainty.

Figures \ref{fig:cofarper} and \ref{fig:svhnper} demonstrate how the quantity of extra unlabeled data influences the effectiveness of our data reduction method, which is designed to prioritize points near the decision boundary. 
For CIFAR-10, the per-batch ratio is set to $0.3$, and \(\beta\) is fixed at $0.4$. For SVHN, the per-batch ratio is set to $0.5$, and \(\beta\) is set to $0.6$. We aim to capture the robust accuracy at the optimal epoch in each case, and we observe that as the amount of included unlabeled data increases, the optimal accuracy is reached at a considerably later epoch.
When the available data is limited, our algorithm excels by selecting a higher proportion of boundary-adjacent data, achieving a balanced selection that enhances robust accuracy. However, as the volume of unlabeled data increases, the algorithm's focus on boundary points can inadvertently lead to overfitting, diminishing the improvement gained in the model's generalization ability. Interestingly, as the data volume continues to grow and more points are added, the performance begins to improve once again, suggesting that the initial overfitting is mitigated by the sheer abundance of data, leading to a recovery in accuracy.

In Figure \ref{fig:ratiocifar}, we present how SSAT's robust accuracy is affected on CIFAR-10 by varying $\gamma$ when $20\%$ extra data from the $1$M generated DDPM samples is used, with \(\beta\) fixed at $0.4$. Similarly, Figure \ref{fig:ratiosvhn} displays the effect of varying the per-batch ratio on robust accuracy for the SVHN dataset, using $10\%$ data from the SVHN extra dataset and fixing \(\beta\) at $0.6$.

\begin{figure}[!t]
\vspace{-0.1in}
    \centering
    \subfloat[CIFAR-10]{
        \centering
        \includegraphics[width=0.45\linewidth]{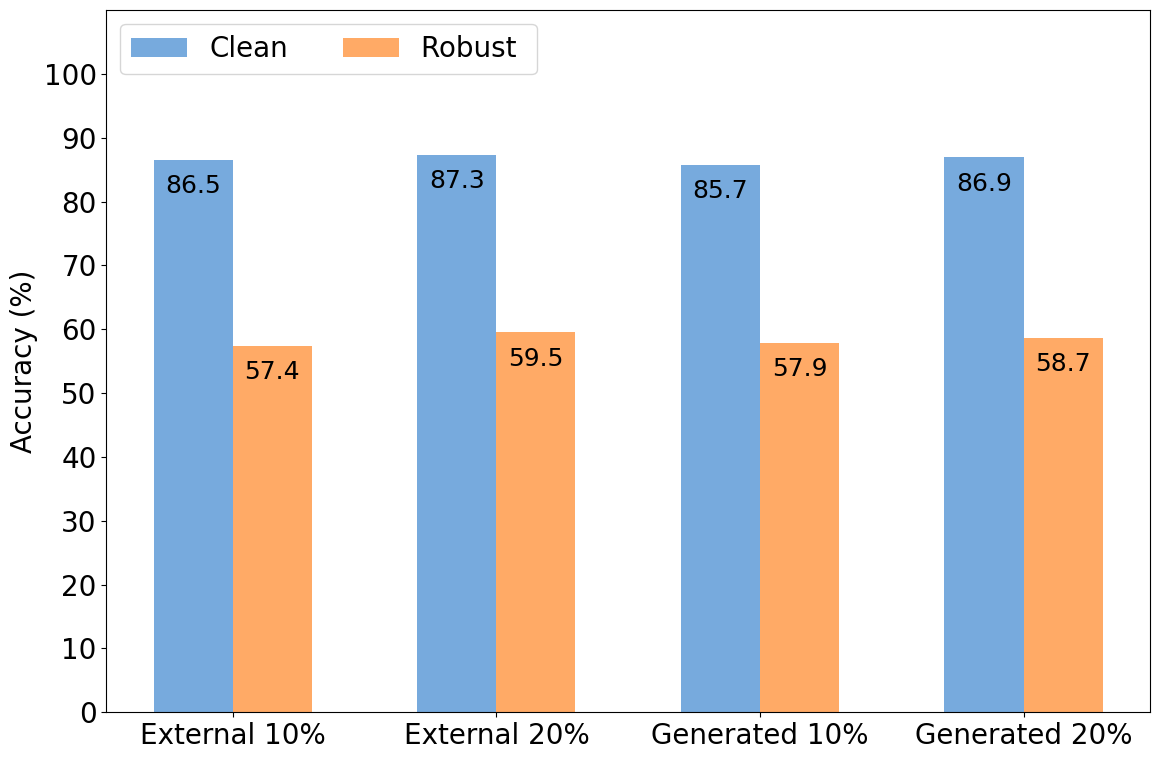}  
        \label{fig:accuracy_cifar10}
    }
    \subfloat[SVHN]{
        \centering
        \includegraphics[width=0.45\linewidth]{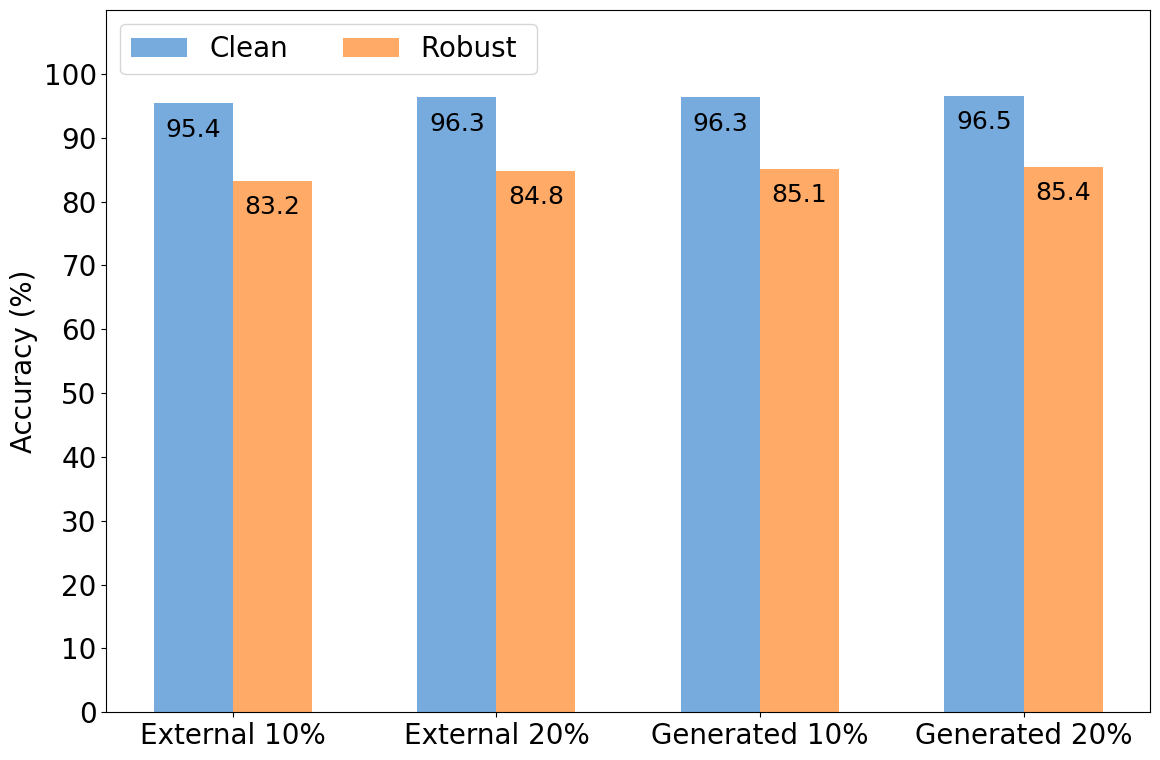}  
        \label{fig:beta_cifar10}
    }
    \caption{Comparisons of SSAT methods with varying amounts of unlabeled data selected using LCS-KM with the CLIP model on (a) CIFAR-10 and (b) SVHN. We report standard and robust accuracies on the labeled test dataset.}
    \vspace{-0.05in}
    \label{fig:comparison_clip}
\end{figure}

\subsection{Pre-Trained CLIP as Intermediate Model}
\label{append:CLIP experiments}

In our experiments, we always began by training an intermediate model $f_{\hat\theta}$ using the same architecture as the final model $\theta_{\mathrm{final}}$. $f_{\hat\theta}$ was trained exclusively on the labeled dataset and then used to generate pseudo labels and select a subset from the extra dataset. However, when dealing with a generated or external dataset that already comes with labels, training an intermediate model specifically for selection undermines the intended benefits of the data selection process, primarily to reduce computational and time complexity. In application scenarios where the size of the labeled dataset $\mathcal{S}_l$ is limited, likely, the intermediate model can not be trained with high accuracy, leading to large amounts of inaccurate pseudo labels.

To address these issues, we investigate the feasibility of utilizing an alternative pre-trained model in LCS for data selection, specifically by leveraging the CLIP model to extract image embeddings \cite{radford2021learning}. CLIP, trained on a vast array of internet images and text, offers a rich and versatile feature space that can be utilized without additional training. By using the capabilities of pre-trained CLIP, we can bypass the intermediate step of training a model from scratch, thereby streamlining the selection process and further enhancing overall efficiency.
This substitution enables us to select a relevant subset without requiring the training of an intermediate model from scratch. Our experiments indicate that using CLIP for data selection yields results that are comparable to those obtained using the trained intermediate model, both in terms of standard and robust accuracies. Figure \ref{fig:comparison_clip} shows the result when using CLIP instead of the previous intermediate model for unlabeled data selection in SSAT.

\begin{table}[!t]
\centering
\caption{Comparison results of SSAT algorithms with varying data selection schemes on SVHN. Instead of pseudo labels, the selected unlabeled data are assigned the ground-truth SVHN labels provided in the dataset.}
\label{tab:SVHN-Result_truelabels}
\vspace{-0.05in}
\small
\centering
\begin{tabular}{l l | c c | c}
    \toprule
    $\bm{\alpha}$ & \textbf{Method} & \textbf{Clean ($\%$)} & \textbf{PGD ($\%$)} & \textbf{\#Epochs}  \\ 
    \midrule 
    $1\%$ & LCS-KM  &  $95.6$ & $82.9$  & $75$ \\ 
    $10\%$ & Random &  $96.3$ & $82.7$ & $75$\\
    $10\%$ & LCS-KM &  $96.7$ & $86.3$  & $75$ \\ 
    $20\%$ & LCS-KM &  $96.4$ & $86.5$  & $75$ \\ 
    \midrule
    $100\%$ & No Selection & $97.5$ &  $86.4$ & $200$ \\ 
    \bottomrule
\end{tabular}
\label{table:SVHN true label}
\end{table}

\begin{table}[t]
\centering
\caption{Comparisons of SSAT performance on CIFAR-10 using the ResNet-18 architecture under various configurations. The model training dataset consists of $50$K labeled CIFAR-10 images and $500$K unlabeled images drawn from Tiny-ImageNet.}
\label{table:CIFAR10-External-resnet}
\vspace{-0.05in}
\begin{tabular}{l l | c c}
    \toprule
    $\bm{\alpha}$ & \textbf{Method} & \textbf{Clean ($\%$)}  & \textbf{PGD ($\%$)}\\
    \midrule
    \multirow{4}{*}{$1\%$} 
    & Random    & $79.8$  & $50.6$\\ 
    & PCS  & $79.4$  & $51.1$\\ 
    & LCS-GMM   & $78.7$  & $51.6$\\ 
    & LCS-KM   & $78.1$  & $51.8$\\ 
    \midrule
    \multirow{4}{*}{$10\%$} 
    & Random    & $82.3$  & $52.0$\\ 
    & PCS  & $82.1$  & $52.6$\\ 
    & LCS-GMM   & $82.0$  & $53.2$\\ 
    & LCS-KM   & $83.3$  & $53.6$\\ 
    \midrule
    \multirow{4}{*}{$20\%$} 
    & Random    & $82.4$  & $53.1$\\ 
    & PCS  & $82.1$  & $53.3$\\ 
    & LCS-GMM   & $82.1$  & $53.7$\\ 
    & LCS-KM   & $83.1$  & $54.0$\\ 
    \midrule
    $100\%$ & No Selection & $83.2$  & $54.7$\\ 
    \bottomrule
\end{tabular}
\end{table}

\begin{figure*}[t] 
    \centering
    \subfloat[TRADES ($\epsilon = 4/255$, PGD-10)]{
        \includegraphics[width=0.23\linewidth]{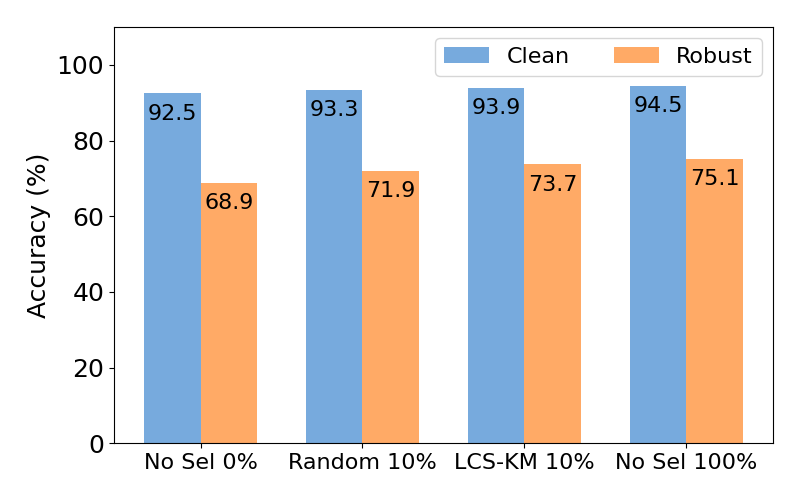}
        \label{fig:4_255}
    }
    \subfloat[TRADES ($\epsilon = 16/255$, PGD-10)]{
        \includegraphics[width=0.23\linewidth]{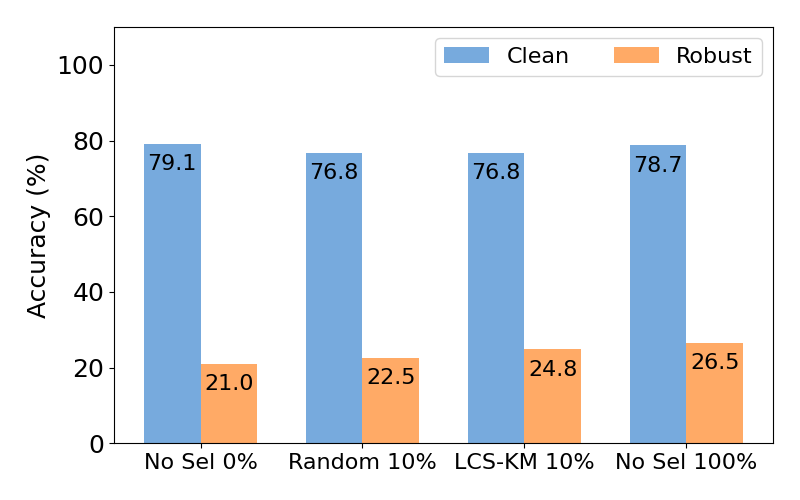}
        \label{fig:16_255}
    }
    \subfloat[AT ($\epsilon = 8/255$, PGD-5)]{
        \includegraphics[width=0.23\linewidth]{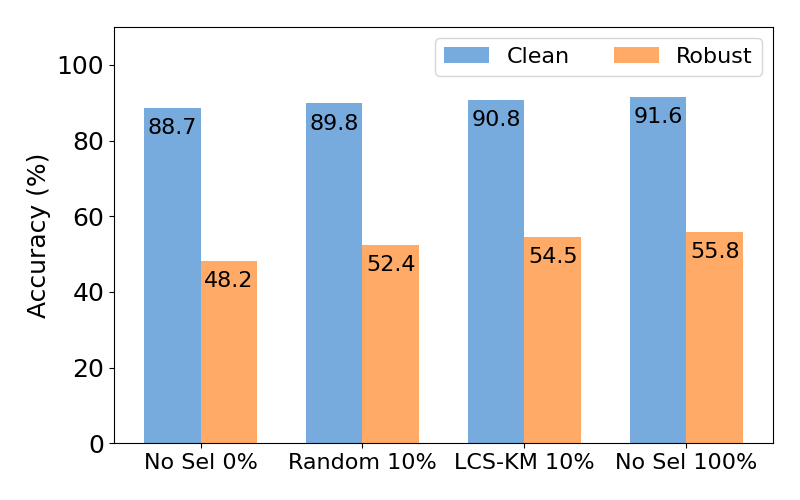}
        \label{fig:pgd-5}
    }
    \subfloat[AT ($\epsilon = 8/255$, PGD-10)]{
        \includegraphics[width=0.23\linewidth]{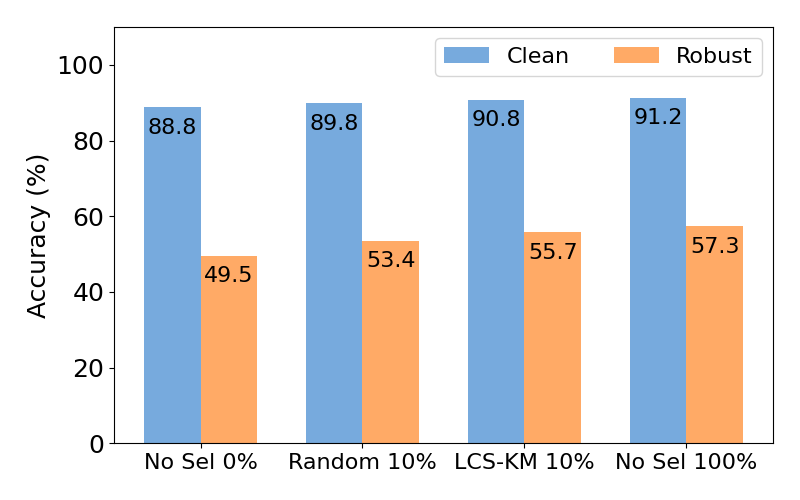}
        \label{fig:-pgd}
    }
    \caption{Illustration of SSAT performance on CIFAR-10 using LCS-KM under $\ell_\infty$ perturbations under various configurations: TRADES using $10$ PGD steps with (a) $\epsilon = 4/255$ and (b) $\epsilon = 16/255$; AT with $\epsilon = 8/255$ using (c) $5$ PGD steps and (d) $10$ PGD steps.}
    \vspace{-0.05in}
    \label{fig:four_graphs_ab}
\end{figure*}

\subsection{SVHN with Ground-Truth Labels}
\label{append:SVHN ground-truth}

We conduct additional experiments to test the performance of our data selection scheme on SVHN with pseudo labels replaced by ground-truth labels for the extra data. Table \ref{table:SVHN true label} shows the comparison results. With ground-truth labels, our LCS schemes can achieve similar robust accuracy by selecting just $10\%$ extra data compared with no selection result.

\subsection{ResNet-18 Model Architecture}
\label{append:resnet 18 additional}

Our main experiments focus on the WideResNet architecture. We further evaluate the performance of SSAT under $\ell_\infty$ perturbations with $\epsilon = 0.031$ using different data selection schemes and varying selection ratios with external unlabeled data on CIFAR-10 using a ResNet-18 model.
Similar to our previous experiments, we train the ResNet-18 model on $50$K labeled samples from CIFAR-10 and a chosen subset of data from $500$K unlabeled images of Tiny-ImageNet.
Our evaluation results are shown in Table \ref{table:CIFAR10-External-resnet}, which suggests that our proposed LCS methods can greatly improve the efficacy of SSAT on ResNet-18 while achieving comparable robustness performance of ``No Selection'' with $\alpha=100\%$. 

\section{Extended Analyses}
\label{app:more_analysis}

In this section, we further evaluate the flexibility of our method across various adversarial training algorithms and perturbation configurations (Section \ref{sec:exploration of generalizability}), and examine the impact of intermediate model quality, particularly under the regime with insufficient labeled data (Section \ref{sec:low-label regime}). 
Without explicit mention, we use our LCS-KM method for selecting critical unlabeled data, as it derives the best-performing model when integrated with SSAT as suggested by Table \ref{table:SVHN+CIFAR10}.

\begin{table*}[!t]
\centering
\caption{Performance of intermediate models trained using different schemes, including self-supervised learning (SSL), pre-trained CLIP, and fully-supervised methods, on CIFAR-10 under varying ratios of labeled data. We also present the corresponding performance of the final model produced by SSAT with LCS-KM, where the ratio of selected unlabeled data $\alpha$ is set at $10\%$.}
\label{tab:intmodel}
\vspace{-0.05in}
\resizebox{0.85\textwidth}{!}{
\begin{tabular}{c | c |  c | c | c}
\toprule
\multirow{2.4}{*}{\textbf{Labeled Ratio}} & \multicolumn{2}{c|}{\textbf{Intermediate Model}} & \multicolumn{2}{c}{\textbf{Final Model (LCS-KM, $\alpha = 10\%$)}} \\
\cmidrule{2-5}
 & \textbf{Training Method} & \textbf{Clean Accuracy ($\%$)} & \textbf{Clean Accuracy ($\%$)} & \textbf{PGD Robust Accuracy ($\%$)} \\ 
\midrule
\multirow{2}{*}{$1\%$}   & SSL ($5\%$ unlabeled)      & $78.8$                  & $82.9$                        & $39.8$                     \\ 
                       & pre-trained CLIP              & $86.8 $                 & $83.9$                        & $40.5$                     \\ \midrule
\multirow{2}{*}{$10\%$}  & SSL ($5\%$ unlabeled)      & $82.9 $                 & $85.0 $                       & $42.4  $                   \\
                       & pre-trained CLIP                 & $86.8 $                 & $85.3$                        & $43.2$                     \\ \midrule
\multirow{2}{*}{$20\%$}  & SSL ($5\%$ unlabeled)       & $87.0$                  & $87.2 $                       & $48.3$                     \\ 
                       & pre-trained CLIP                 & $86.8$                  & $85.6$                        & $47.7$                     \\ \midrule
\multirow{2}{*}{100\%} & fully-supervised      & $90.5$                  & $86.2$                        & $58.2 $                    \\
                       & pre-trained CLIP               & $86.8$                  & $86.5$                        & $57.4$                     \\ \bottomrule
\end{tabular}
}
\end{table*}

\subsection{Further Exploration of Generalizability}
\label{sec:exploration of generalizability}

We compare four different settings of SSAT on CIFAR-10: random selection with $\alpha$ ratio of unlabeled data, selecting $\alpha$ ratio using LCS-KM, using only labeled data ($\alpha = 0\%$), and using the entire dataset ($\alpha = 100\%$).

\shortsection{Perturbation Size} 
We consider external data from TinyImages as the unlabeled dataset.
We investigate the effect of varying the $\ell_\infty$ perturbation size $\epsilon$, setting it to $4/255$ and $16/255$ while keeping TRADES as the adversarial training algorithm, similar to the experiments in Table~\ref{table:SVHN+CIFAR10}. The results are depicted in Figures~\ref{fig:4_255} and~\ref{fig:16_255}, suggesting that our LCS-KM method is highly effective compared to random selection and no selection, approaching the robust accuracies achieved by fully utilizing unlabeled data.
The trends are consistent across varying perturbation sizes.

\shortsection{Adversarial Training} 
We explore the impact of replacing TRADES with vanilla AT as the training algorithm.
Similarly, we use external data from TinyImages as the unlabeled dataset.
The results for PGD-based AT are shown in Figure~\ref{fig:-pgd}. Additionally, we reduce the number of PGD steps from $10$ to $5$, and the outcomes are summarized in Figure~\ref{fig:pgd-5}.  
Across all configurations, our LCS-KM selection consistently outperforms random selection and achieves accuracy comparable to using the entire labeled dataset, suggesting the generalizability and robustness of our methods.

\begin{figure}[!t] 
\vspace{-0.1in}
    \centering
    \subfloat[External data]{
        \includegraphics[width=0.45\linewidth]{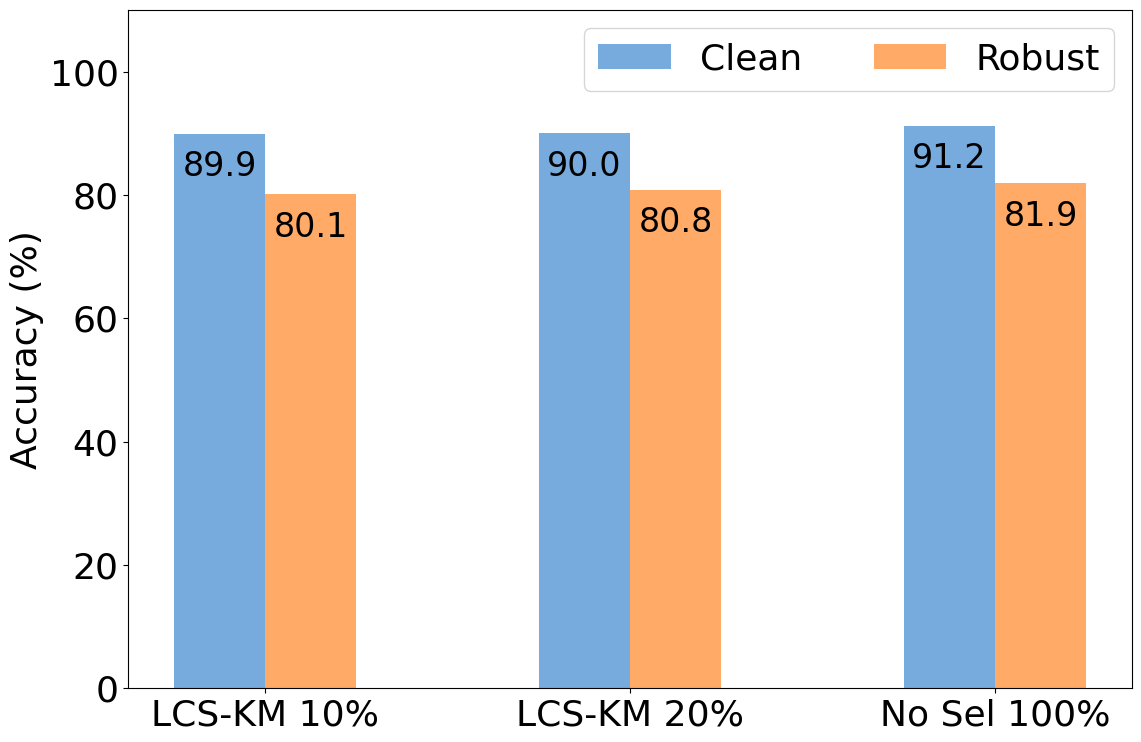}
        \label{fig:externall2}
    }
    \subfloat[Generated data]{
        \includegraphics[width=0.45\linewidth]{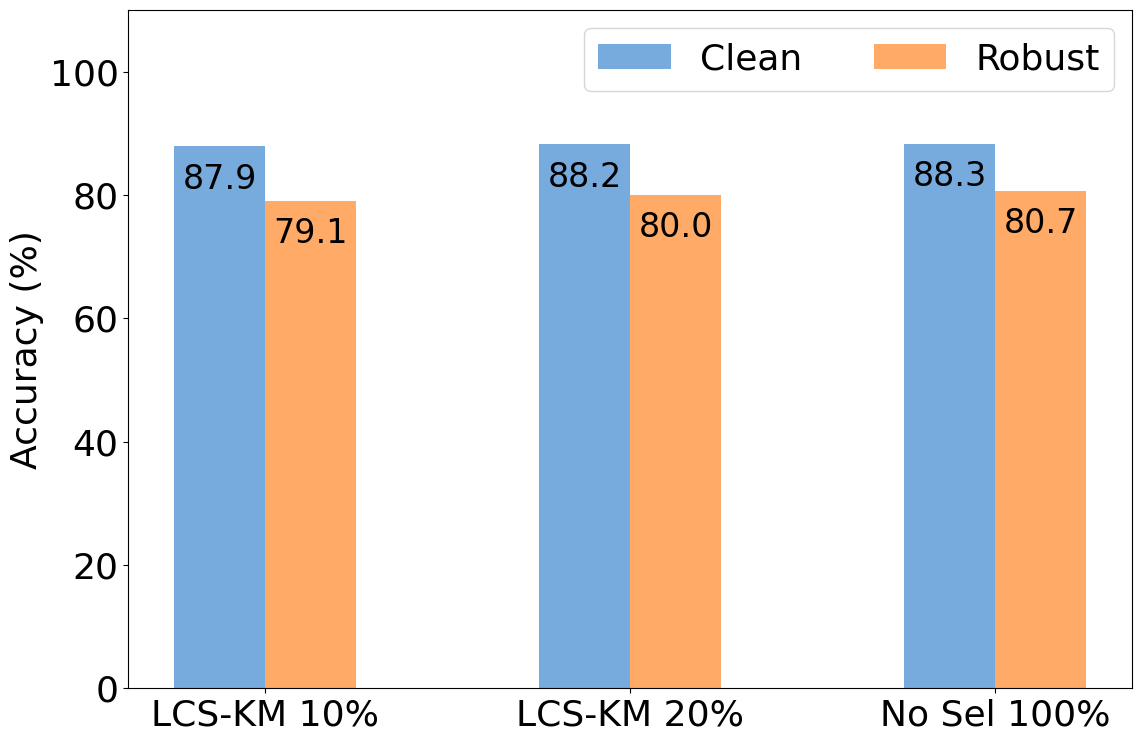}
        \label{fig:generatedl2}
    }
    \caption{Illustration of SSAT performance on CIFAR-10 under $\ell_2$ perturbations with $\epsilon = 128/255$ using LCS-KM by varying $\alpha$ with: (a) $500$K external TinyImages and (b) $1$M DDPM-generated data.}
    \vspace{-0.1in}
    \label{fig:two_graphs_ab}
\end{figure}

\shortsection{$\bm\ell_2$ Perturbations}
We further evaluate the performance of LCS-KM under $\ell_2$ perturbations with size $\epsilon = 128/255$. In this experiment, we augment CIFAR-10 with either external unlabeled data sourced from Tiny-ImageNet or synthetic data generated via DDPM.
We consider three configurations of selecting $10\%$, $20\%$, and $100\%$ extra unlabeled data. Our findings are presented in Figures \ref{fig:externall2} and \ref{fig:generatedl2}, which reveal that selecting $20\%$ unlabeled data achieves nearly equivalent robust accuracy compared to using the full dataset with $\alpha=100\%$. These results again confirm the efficiency and effectiveness of LCS-KM when adopted in SSAT algorithms.

\subsection{Low Labeled Data Regime}
\label{sec:low-label regime}

We further investigate how the choice of intermediate model and the amount of labeled data affect the final performance. For all experiments, we use a subset of the CIFAR-10 dataset as labeled data, supplemented with $10\%$ of data selected from a pool of $500$K unlabeled images sourced from Tiny-ImageNet. This additional data is selected using the intermediate model. The amount of unlabeled data remains constant across experiments, while the quantity of labeled data is varied.
The intermediate model is trained using self-supervised learning (SSL) with a $5\%$ randomly selected subset of the unlabeled dataset, provided the labeled data ratio is low. 
In addition, we include a pre-trained CLIP model \cite{radford2021learning} as an alternative intermediate model for comparison. 

Table~\ref{tab:intmodel} summarizes our results.
The quality of the intermediate model significantly impacts the robust accuracy of the final model. In low-label regimes (e.g., $1\%$ or $10\%$), models trained using our intermediate model yield comparatively lower robust accuracy than those leveraging the pre-trained CLIP model. However, as the amount of labeled data for the intermediate model increases (e.g., $20\%$ or $100\%$), the performance of the final model becomes similar to or better than that of its counterparts based on the CLIP intermediate model. This trend demonstrates the effect of the intermediate model on the final robust accuracy.
These observations highlight the importance of having a robust intermediate model for achieving strong performance. Additionally, we observe that pre-trained models, such as CLIP, can be a good alternative to training an intermediate model, especially in low-labeled-data regimes. 
Detailed results and discussions on the pre-trained CLIP intermediate model are provided in Appendix~\ref{append:CLIP experiments}.

\begin{table}[!t]
\centering
\caption{Full version of Table \ref{tab:cifar10_finetuning}. We evaluate full SSAT ($\alpha=100\%$) on CIFAR-10 under $\ell_\infty$ perturbations with $\epsilon=0.031$ under various learning rate schedulers with/without early stopping.}
\label{tab:early_stop_cifar10 full}
\vspace{-0.05in}
\small
\resizebox{0.5\textwidth}{!}{
\begin{tabular}{l l c | c c c}
    \toprule
    \textbf{Setting} & \textbf{Schedule} & \textbf{Epoch} & \textbf{Clean} ($\%)$ & \textbf{PGD} ($\%)$ & \textbf{AA} ($\%)$ \\
    \midrule
    \multirow{4}{*}{External} 
    & \multirow{2}{*}{Cosine} & $100$ & $84.1$ & $54.0$ & $50.4$ \\
    &  & $200$ & $89.3$ & $62.2$ & $59.8$ \\
    & \multirow{2}{*}{Stepwise} & $100$ & $88.0$ & $60.6$ & $57.6$ \\
    &  & $200$ & $89.7$ & $62.5$ & $58.6$ \\
    \midrule
    \multirow{4}{*}{DDPM} 
    & \multirow{2}{*}{Cosine} & $100$ & $78.6$ & $52.1$ & $48.6$ \\
    &  & $400$ & $86.2$ & $62.4$ & $59.9$ \\
    & \multirow{2}{*}{Stepwise} & $100$ & $83.0$ & $58.5$ & $55.9$ \\
    &  & $400$ & $85.7$ & $61.8$ & $58.4$ \\
    \bottomrule
\end{tabular}
}
\vspace{-0.1in}
\end{table}

\subsection{Performance of Full SSAT Stopped at Earlier Epochs}
\label{app:full SSAT early stop}

To investigate whether early stopping can serve as an effective strategy for full SSAT rather than training for the complete number of epochs, we evaluate SSAT models on CIFAR-10 using $100\%$ of the unlabeled data with two popular learning rate scheduling schemes, both with an initial learning rate of $0.1$: (Cosine) cosine annealing and (Stepwise) stepwise decay with a decaying factor of $10$ at training epochs $75$ and $90$. We compare the performance of SSAT stopped at epoch $100$, the same as our data reduction methods, against SSAT models trained for the full duration: $200$ epochs for $500$K external TinyImages and $400$ epochs for $1$M DDPM-generated data. 
Table~\ref{tab:early_stop_cifar10 full} presents the complete results. Below, we summarize the key observations regarding the interplay between learning rate scheduling and the training duration of SSAT:

\shortsection{Cosine vs. Stepwise}
The two learning rate schedulers exhibit distinct training dynamics during the early epochs. At epoch $100$, models trained with cosine annealing show significantly lower performance (e.g., $50.4\%$ AA robust accuracy for external data and $48.6\%$ for DDPM-generated data) compared to stepwise decay ($57.6\%$ and $55.9\%$, respectively). 
However, when trained to completion, cosine annealing achieves more competitive final robustness performance than stepwise scheduling, approximately $1\%-1.5\%$ improvement in robust accuracy when evaluated using AutoAttack.

\shortsection{Effectiveness of Early Stopping}
For the stepwise scheduler, early stopping at epoch $100$ can achieve robust accuracy not far from the robustness of the final model returned by full SSAT. For external data, the robustness gap between epochs $100$ and $200$ is only $1\%$ with a doubled training time. Similarly, for DDPM data, early stopping at epoch $100$ yields $55.9\%$ AA robust accuracy compared to $58.4\%$ at epoch $400$, a difference of $2.5\%$ with $4\times$ more training. 
In comparison, early stopping is not effective for full SSAT when using the cosine annealing scheduler.
Prolonged training of full SSAT algorithms from epoch $100$ gradually increases the clean and robust accuracy performance until reaching peak performance, regardless of the type of learning rate scheduler implemented. These observations are aligned with the full SSAT curves illustrated in Figure \ref{fig:100perc}, suggesting that learning is not complete for SSAT with $\alpha=100\%$ if it is simply stopped at an early epoch.

\shortsection{Comparison with LCS-KM}
Even with the stepwise scheduler and early stopping employed, full SSAT  early stopped at epoch $100$ achieves AA robust accuracy of $57.6\%$ (external) and $55.9\%$ (DDPM-generated), slightly lower than our LCS-KM method with $20\%$ unlabeled data that achieves $57.8\%$ and $57.2\%$ for the respective settings (Table~\ref{table:SVHN+CIFAR10}). Under cosine scheduling, the robustness improvement of using LCS-KM is much more pronounced compared to full SSAT with early stopping.
These results demonstrate the advantages of our data reduction methods, offering a viable solution to improve the data efficiency of SSAT algorithms and accelerate the learning process while largely preserving their robustness advantages.

\subsection{Further Discussions}
\label{append:further discussions}

Our work introduced several strategies to improve the data efficiency of SSAT algorithms by prioritizing unlabeled data adjacent to the model's decision boundaries. However, the success of our methods hinges on the hypothesis that the intermediate model’s latent representations are reliable for pseudo-labeling and prioritizing unlabeled data, thereby improving robust generalization, which may falter in scenarios where the model is undertrained or poorly calibrated. In addition, while our methods effectively reduce the volume of unlabeled data, their final performance can be sensitive to hyperparameter configurations, such as the balance between boundary and non-boundary points. 
Developing automatic tools to automate the process of hyperparameter selection, as well as conducting in-depth theoretical studies to fully comprehend how these strategies influence decision boundary geometry and the model's capacity for robust generalization, along with their adaptability to potential variations, would be meaningful future directions.

Moreover, reducing the overall runtime required for adversarial training without sacrificing robustness is critical for achieving efficient and scalable robust learning.  
In the prior literature, two complementary strategies have been proposed:  
(i) decreasing the computation of each training epoch, such as free and fast adversarial training~\cite{shafahi2019adversarial, wong2020fast}, and other more efficient optimizers for generating perturbations in an adversarial training framework~\cite{chen2022efficient, zhao2023fast}, and  
(ii) reducing the total number of epochs needed for convergence, such as implementing early stopping~\cite{rice2020overfitting}. 
While the former techniques can reduce the per-epoch runtime, the overall training process is less stable compared to vanilla adversarial training, where PGD with sufficient iterations is used to solve the inner maximization problem. 
Our work falls into the second category, where we adopt early stopping to accelerate the convergence of SSAT algorithms.
Nevertheless, developing effective strategies based on our data reduction schemes and more efficient inner maximization solvers represents a promising future direction for further improving the practicality of SSAT algorithms.

\end{document}